\documentclass{article}

\PassOptionsToPackage{sort&compress,square,comma,numbers}{natbib}
 \usepackage[final]{nips_2017}

\usepackage[utf8]{inputenc} 
\usepackage[T1]{fontenc}    
\usepackage{hyperref}       
\usepackage{url}            
\usepackage{booktabs}       
\usepackage{amsfonts}       
\usepackage{nicefrac}       
\usepackage{microtype}      

\usepackage{graphicx} 
\usepackage{floatrow}
\usepackage{subfig}  
   \graphicspath{{figs/}}
   \DeclareGraphicsExtensions{.pdf,.jpg,.png,.eps}

\usepackage{siunitx}
\usepackage{multirow}

\usepackage{amsmath,amssymb}
\usepackage{xspace}
\usepackage{wrapfig,caption}

\newcolumntype{L}{S[table-format=2.1]@{\hskip 10pt}}
\newcolumntype{R}{S[table-format=2.1]}
\newcolumntype{?}{!{\vrule}}
\newcommand{\mytimes}{\medmuskip=0mu\times}
\newcommand{\dense}{\thickmuskip=2mu}

\newcommand{\secref}[1]{Section~\ref{#1}}
\renewcommand{\eqref}[1]{(\ref{#1})}
\newcommand{\figref}[1]{Fig.~\ref{#1}}

\newcommand{\tabref}[1]{Table~\ref{#1}}

\newcommand{\ie}{\textrm{i.e.}\xspace}
\newcommand{\eg}{\textrm{e.g.}\xspace}

\newenvironment{eq}
  {\vspace*{-2pt}\equation}{\vspace*{-2pt}\endequation}
\newcommand{\sss}{\scriptscriptstyle}

\newcommand{\mean}{\ensuremath{\mathbb{E}}\xspace}
\newcommand{\model}{\ensuremath{M}\xspace}
\newcommand{\param}{\ensuremath{\boldsymbol{\theta}}\xspace}
\newcommand{\paramset}{\ensuremath{\Theta}\xspace}
\newcommand{\task}{\ensuremath{W}\xspace}
\newcommand{\taskset}{\ensuremath{\mathcal{W}}\xspace}
\newcommand{\taskspace}[1]{\ensuremath{\underline{#1}}}
\newcommand{\modelspace}[1]{\ensuremath{#1}}

\def\abovestrut#1{\rule[0in]{0in}{#1}\ignorespaces}

\def\abovespace{\abovestrut{0.15in}}

\title{QMDP-Net: Deep Learning for Planning under Partial Observability}
\author{
  Peter Karkus$^{1,2}$
  \qquad
  David Hsu$^{1,2}$
  \qquad
  Wee Sun Lee$^2$  \vspace{0.25cm}\\
  $^1$NUS Graduate School for Integrative Sciences and Engineering  \\
  $^2$School of Computing\vspace{0.25cm}\\
  National University of Singapore\\
  \texttt{\{karkus, dyhsu, leews\}@comp.nus.edu.sg} \\
}

\begin{document}

\maketitle


\begin{abstract} \itshape\small This paper introduces the \emph{QMDP-net}, a
  neural network architecture for planning under partial observability.  The
  QMDP-net combines the strengths of model-free learning and model-based
  planning. It is a recurrent policy network, but it represents a policy for
  a parameterized set of tasks by
  connecting a model with a planning algorithm that solves the model, thus
  embedding the solution structure of planning in a network learning
  architecture. The QMDP-net is fully differentiable and allows for end-to-end
  training.  We train a QMDP-net on different tasks so that it can generalize
  to new  ones  in the parameterized task
  set  and ``transfer'' to other similar tasks beyond the set.
  In preliminary experiments, QMDP-net showed strong performance on
  several robotic tasks in simulation.  Interestingly, while QMDP-net
  encodes the QMDP algorithm, it sometimes outperforms the QMDP algorithm in
  the experiments, as a result of   end-to-end learning.
\end{abstract}


\section{Introduction}\label{sec:intro}
\begin{figure*}[!t]
  \centering
   \sidesubfloat[][]{ \hspace{-0.27cm}\includegraphics[height=3.4cm]{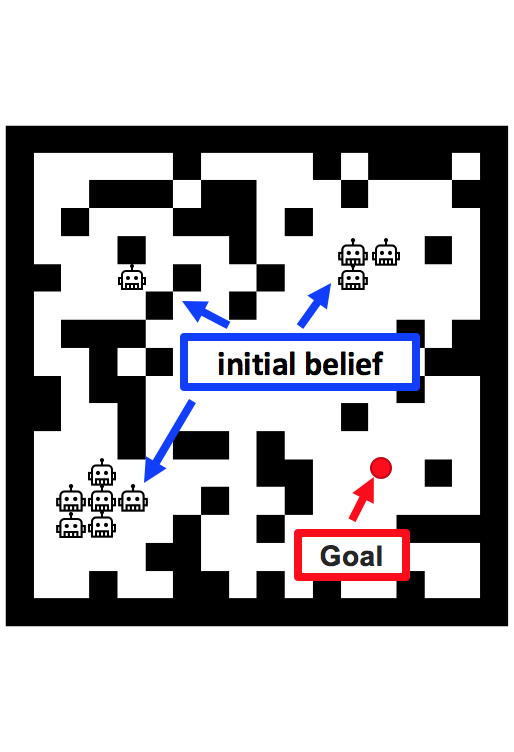}\label{fig:fig1a} \hspace{0.14cm}}
   \sidesubfloat[][]{ \hspace{-0.27cm}\includegraphics[height=3.4cm]{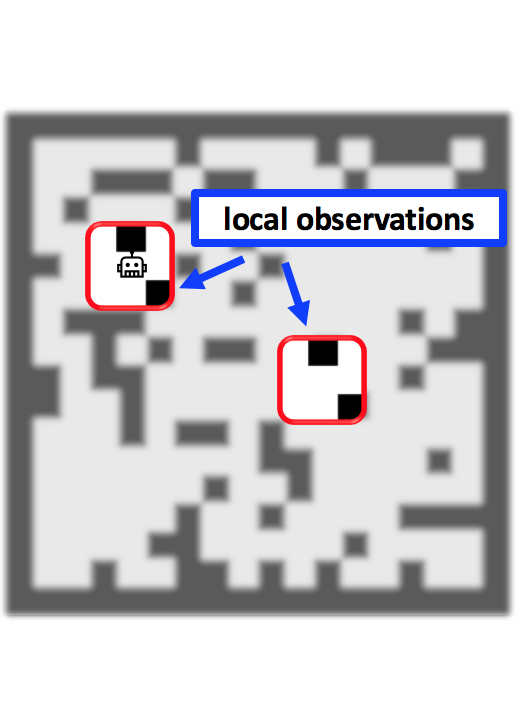}\label{fig:fig1b} \hspace{0.14cm}} 
   \sidesubfloat[][]{ \hspace{-0.15cm}\includegraphics[height=3.4cm]{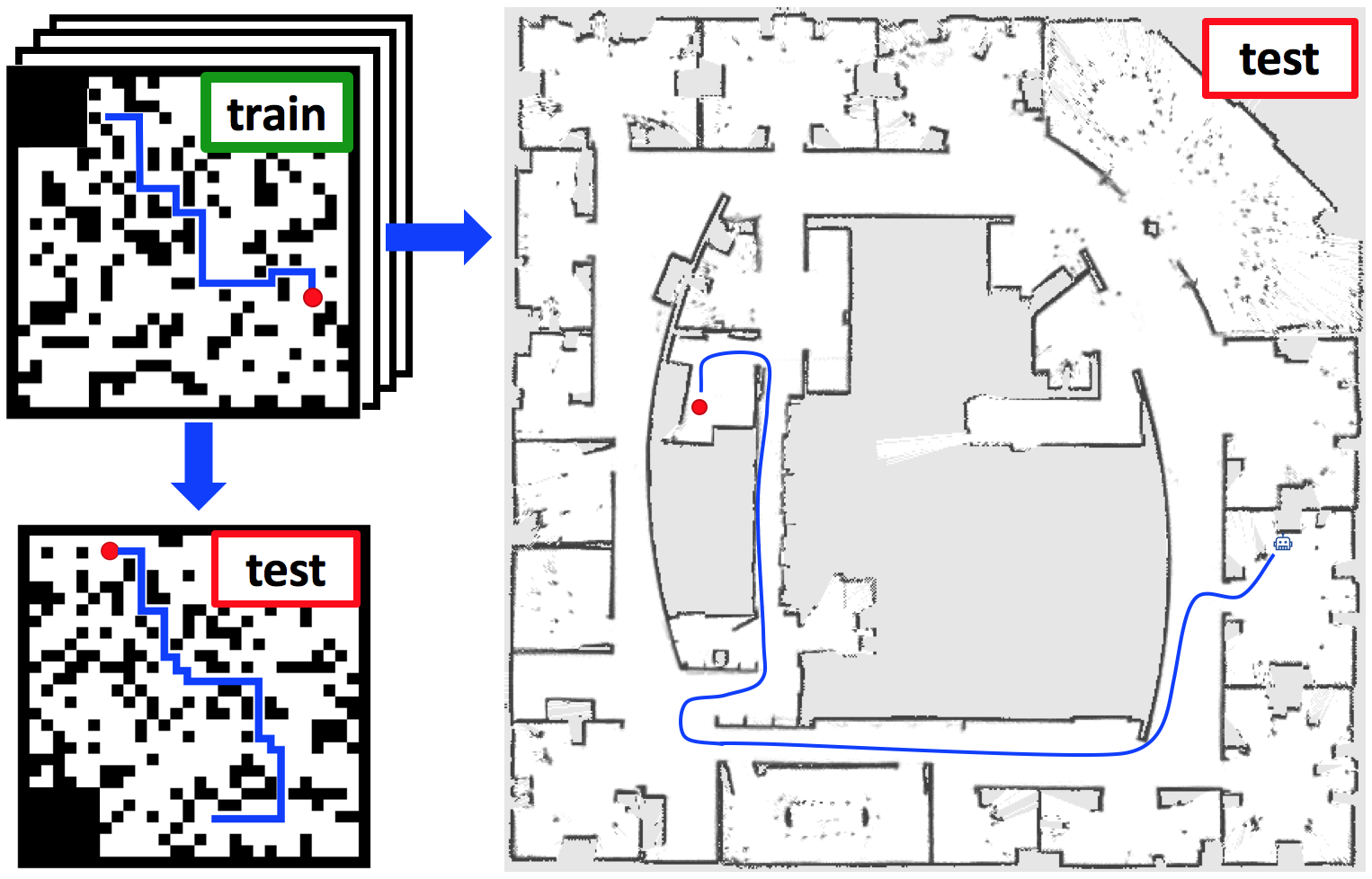}\label{fig:fig1c} \hspace{0.1cm}} 
   \sidesubfloat[][]{ \hspace{-0.15cm}\includegraphics[height=3.4cm]{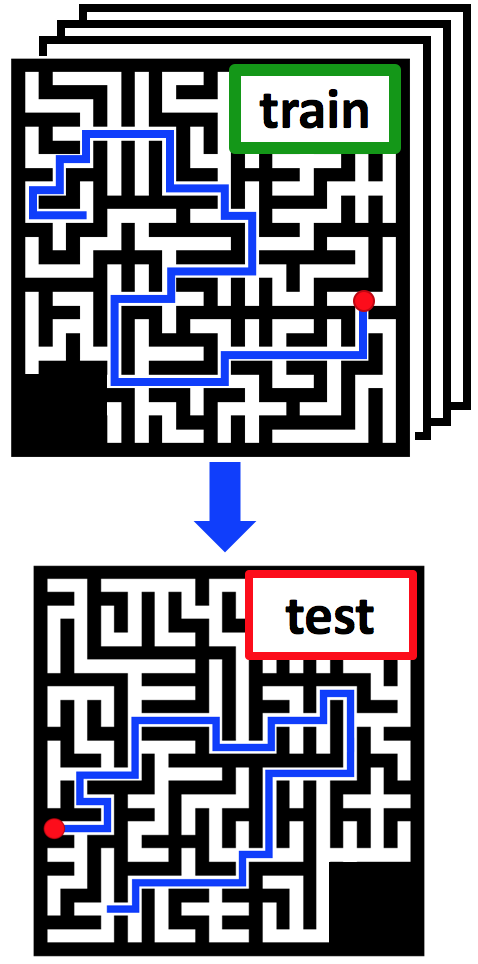}\label{fig:fig1d} } 
   \caption{A robot learning to navigate in partially observable grid worlds.
     (a) The robot has a map. It has a belief over the initial state, but
     does not  know the exact initial state.  (b)
     Local observations are ambiguous and
     are insufficient to determine the exact state. (c, d) A policy trained
     on expert demonstrations in a set of randomly generated
     environments generalizes to a
     new environment. It also ``transfers'' to
     a much larger real-life environment, represented as a LIDAR
     map~\cite{radish}.}
\label{fig:fig1}
\end{figure*}

Decision-making under uncertainty is of fundamental importance, but it
is computationally hard, especially under partial
observability~\cite{papadimitriou1987complexity}.
In a partially observable world, the agent cannot determine the
state exactly based on the current  observation;
 to plan  optimal
actions, it must integrate information over
the past history of actions and  observations.
See \figref{fig:fig1} for an example.
 In the model-based
approach, we may formulate the problem as a \emph{partially
  observable Markov decision process} (POMDP).
Solving POMDPs exactly is computationally intractable in the worst case~\cite{papadimitriou1987complexity}.
Approximate POMDP algorithms
have made dramatic progress on solving large-scale POMDPs~\cite{pineau2003applying,spaan2005perseus,kurniawati2008sarsop,silver2010monte,ye2017despot};
however,  manually constructing POMDP models or learning them  from data
remains difficult. In the model-free approach, we
directly search for an optimal solution within a
policy class. If we do not restrict the policy class, the difficulty is
data and computational efficiency. We may choose a
parameterized policy class. The effectiveness of policy search is then
constrained by this a priori choice.

Deep neural networks have brought unprecedented success in many
domains~\cite{krizhevsky2012imagenet, mnih2015human, silver2016mastering} and
provide a distinct new approach to decision-making under uncertainty.  The
deep Q-network (DQN), which consists of 
a convolutional neural network (CNN) together
with a fully connected layer, has successfully tackled many Atari games with
complex visual input~\cite{mnih2015human}.
Replacing the
post-convolutional
fully connected layer of DQN by a recurrent
LSTM layer allows it to deal with partial
observaiblity~\cite{hausknecht2015deep}. However, compared with
planning, this approach fails 
to exploit the underlying sequential nature of decision-making. 

We introduce \emph{QMDP-net}, a neural network architecture for planning under
partial observability.  QMDP-net combines the strengths of model-free learning
and
model-based planning. A QMDP-net is a recurrent policy network, but it
represents a policy by connecting a POMDP model with an algorithm that solves
the model, thus embedding the solution structure of planning in a network
learning
architecture. Specifically, our network uses QMDP~\cite{littman1995learning},
a simple, but fast approximate POMDP algorithm, though other more
sophisticated POMDP algorithms could be used as well.

A QMDP-net consists of two main network
modules (\figref{fig:rnn}). One represents a Bayesian filter, which integrates
the history of an agent's actions and observations into a \emph{belief}, \ie{}
a probabilistic estimate of the agent's state.  The other represents the QMDP
algorithm, which chooses the action given the current belief. Both modules
are differentiable, allowing the entire network to be trained end-to-end.

\floatsetup[figure]{style=plain,subcapbesideposition=top}

We train a QMDP-net on  expert demonstrations in a set of randomly generated
environments. The trained policy generalizes to new environments and also
``transfers'' to more complex environments (\figref{fig:fig1}c--d).  Preliminary experiments show that QMDP-net outperformed
state-of-the-art network architectures on several robotic tasks in simulation.
It successfully solved difficult POMDPs that require reasoning over many time
steps, such as the well-known Hallway2 domain~\cite{littman1995learning}.
Interestingly, while QMDP-net encodes the QMDP algorithm, 
it sometimes outperformed the QMDP algorithm in our experiments,
as a result of  end-to-end learning.

\section{Background}\label{sec:related}
\subsection{Planning under Uncertainty}
%

A POMDP is formally defined as a tuple $(S,A,O,T,Z,R)$, where
$S$, $A$ and $O$ are the state, action, and observation space,
respectively. The state-transition function $T(s, a, s') = P(s' | s, a)$ defines the
probability of the agent being in state $s'$ after taking action $a$ in state
$s$. The observation function $Z(s, a, o) = p(o | s, a)$ defines the
probability of receiving observation $o$ after taking action $a$ in
state~$s$. The reward function $R(s,a)$ defines the immediate reward for taking
action $a$ in state $s$.

In a partially observable world, the agent does not know its exact state. It
maintains a \emph{belief}, which is a probability distribution over $S$.
The agent starts with an initial belief $b_0$ and updates the belief $b_t$
at each time step $t$  with a Bayesian filter:  
\begin{eq}
\textstyle b_t(s') = \tau(b_{t-1},a_t,o_t) 
= \eta O(s',a_t,o_t)
\sum_{s \in S}T(s,a_t,s')b_{t-1}(s),
\end{eq}
where $\eta$ is a normalizing constant.
The belief $b_t$ recursively integrates information from the \emph{entire} past
history  $(a_1, o_1, a_2, o_2, \ldots, a_t, o_t)$ for decision making.
POMDP planning  seeks a \emph{policy} $\pi$ that maximizes
the \emph{value}, \ie, the expected total discounted reward:
\begin{eq}
\textstyle V_{\pi}(b_0) = \mean \bigl( \sum_{t=0}^{\infty} \gamma^t R(s_t, a_{t+1})
\;\bigl|\; b_0, \pi
\bigr),
\end{eq}
where $s_t$ is the state at time $t$, $a_{t+1} = \pi(b_t)$ is the action
that the policy $\pi$ chooses
at time $t$, and $\gamma \in (0,1)$ is a discount
factor.

\subsection{Related Work}

To learn policies for decision making in partially observable domains, one
approach is to learn models \cite{littman2002predictive, shani2005model,
  boots2011closing} and solve the models through planning.  An alternative is
to learn policies directly \cite{baxter2001infinite,bagnell2003policy}. Model
learning is usually not end-to-end. While policy learning can be end-to-end,
it does not exploit model information for effective generalization.  Our
proposed approach combines model-based and model-free learning by embedding a
model and a planning algorithm in a recurrent neural
network (RNN) that represents a policy and then training the network end-to-end.

RNNs have been used earlier for learning in partially observable domains
\cite{hochreiter1997long, bakker2003robot, hausknecht2015deep}.  In
particular, Hausknecht and Stone extended DQN~\cite{mnih2015human}, a
convolutional neural network (CNN), by replacing its post-convolutional fully
connected layer with a recurrent LSTM layer~\cite{hausknecht2015deep}.
Similarly, \citet{mirowski2016learning} considered learning to navigate in
partially observable 3-D mazes. The learned policy generalizes over different
goals, but in a fixed environment.  Instead of using the generic LSTM, our
approach embeds algorithmic
structure specific to sequential decision making in the network
architecture and aims to learn a policy that generalizes to new environments.

The idea of embedding specific computation structures in the neural network
architecture has been gaining attention recently.
Tamar et al.  implemented value iteration in a neural network, called Value
Iteration Network (VIN), to
solve Markov decision processes (MDPs) in fully observable domains, where an
agent knows its exact state and does not require
filtering~\cite{tamar2016value}.
Okada et al. addressed a related problem of path integral optimal control,
which allows for continuous states and actions~\cite{OkaRig17}. 
Neither addresses the issue of partial
observability, which drastically increases the computational complexity of
decision making~\cite{papadimitriou1987complexity}.
\citet{haarnoja2016backprop} and \citet{jonschkowski2016} developed end-to-end
trainable Bayesian filters for probabilistic state estimation.  Silver et
al. introduced Predictron for value estimation in Markov reward
processes~\cite{silver2016predictron}.  They do not deal with decision making
or planning.  Both \citet{shankar2016reinforcement} and
\citet{gupta2017cognitive} addressed planning under partial observability.
The former focuses on learning a model rather than a policy. The learned model
is trained on a fixed environment and does not generalize to new ones.  The
latter proposes a network learning approach to robot navigation in an unknown
environment, with a focus on mapping.  Its network architecture contains a
hierarchical extension of VIN for planning and thus does not deal with partial
observability during planning.
The QMDP-net extends the prior work on network architectures for
MDP planning and for Bayesian filtering. 
It imposes the POMDP model and computation structure priors on the
entire network architecture for planning
under  partial observability.

\section{Overview}\label{sec:overview}

We want to learn a policy that enables an agent to act effectively in a
diverse set of partially observable stochastic environments.  Consider, for
example, the robot navigation domain in \figref{fig:fig1}.  The environments
may correspond to different buildings.  The robot agent does not observe its
own location directly, but estimates it based on noisy readings from a laser
range finder.  It has access to building maps, but does not have models of
its own dynamics and sensors.  While the buildings may differ significantly in
their layouts, the underlying reasoning required for effective navigation is
similar in all buildings.  After training the robot in a few buildings, we
want to place the robot in a new building and have it navigate effectively to
a specified goal.

Formally, the agent learns a policy for a parameterized set of tasks in
partially observable stochastic environments:
$\taskset_{\paramset} = \{\task(\param) \mid \param\in\paramset \}$,
where \paramset is the set of all parameter values.  The parameter
value~$\param$ captures a wide variety of task characteristics that vary 
within the set, including
environments, goals, and agents.  In our robot navigation example,
\param encodes a map of the environment, a goal, and a belief over the robot's
initial state.  We assume that all tasks in $\taskset_{\paramset}$ share the
same state space, action space, and observation space.
The agent does not have prior models of its own
dynamics, sensors, or task objectives.  After training on tasks for some
subset of values in \paramset, the agent learns a policy that solves
$\task(\param)$ for any given $\param\in \paramset$.

A key issue is a general representation of a policy for
$\taskset_{\Theta}$, without knowing the specifics of $\taskset_{\Theta}$
or its parametrization.
We introduce the QMDP-net, a recurrent policy network.
A QMDP-net represents a policy by connecting a parameterized POMDP model with
an approximate POMDP algorithm and embedding both in a single, differentiable
neural network.  Embedding the model allows the policy to generalize over
$\taskset_\Theta$ effectively.  Embedding the algorithm allows us to train the
entire network end-to-end and learn a  model that compensates for the
limitations of the approximate algorithm.

Let
$\dense \model(\param) = (S, A, O, f_{\sss T}(\cdot | \param), f_{\sss Z}(\cdot
| \param), f_{\sss R}(\cdot | \param))$ be the embedded POMDP model,
where $S, A$ and $O$ are the shared state space, action space, observation space
designed manually for all tasks in  $\taskset_\paramset$ and
$f_{\sss T}(\cdot |\cdot), f_{\sss Z}(\cdot | \cdot), f_{\sss R}(\cdot |
\cdot)$ are the state-transition, observation, and reward functions to
be learned from data.
It may appear that
a perfect answer to our learning problem would 
have $f_{\sss T}(\cdot | \param), f_{\sss Z}(\cdot | \param), \text{ and } f_{\sss R}(\cdot | \param)$ represent the ``true''
underlying models of dynamics, observation, and reward for the task
$\task(\param)$.  This is true only if the embedded
POMDP algorithm is exact, but not true in general. The agent may learn an
alternative model to mitigate an approximate algorithm's limitations and obtain
an overall better policy. In this sense, while QMDP-net embeds a POMDP model
in the network architecture, it aims to learn a good policy rather than a
``correct'' model.

\begin{figure*}[t]
  \centering
   \sidesubfloat[][]{ \includegraphics[height=3.0cm]{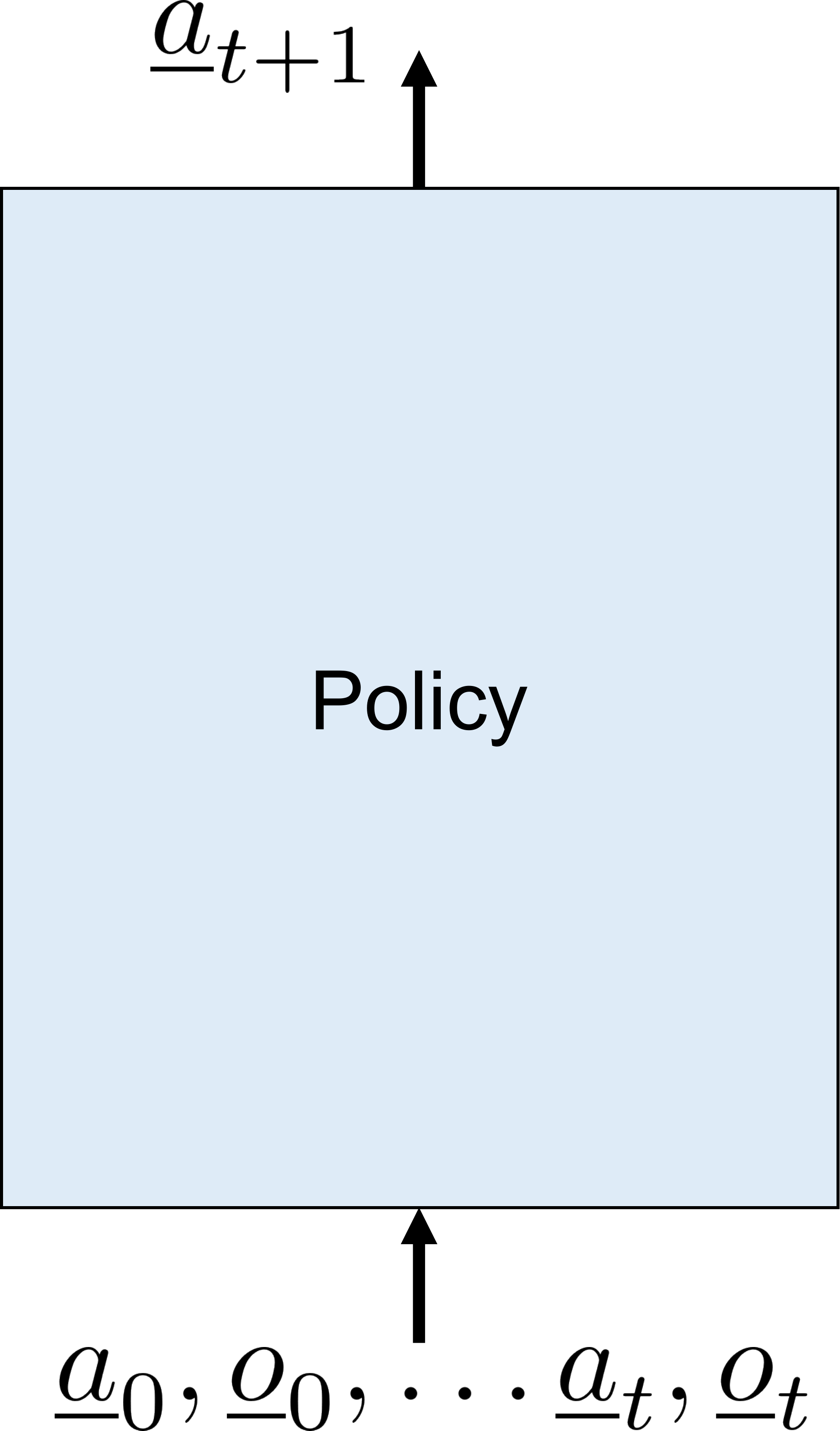}\label{fig:rnn1} }\hspace{3.2mm} 
   \sidesubfloat[][]{ \includegraphics[height=3.0cm]{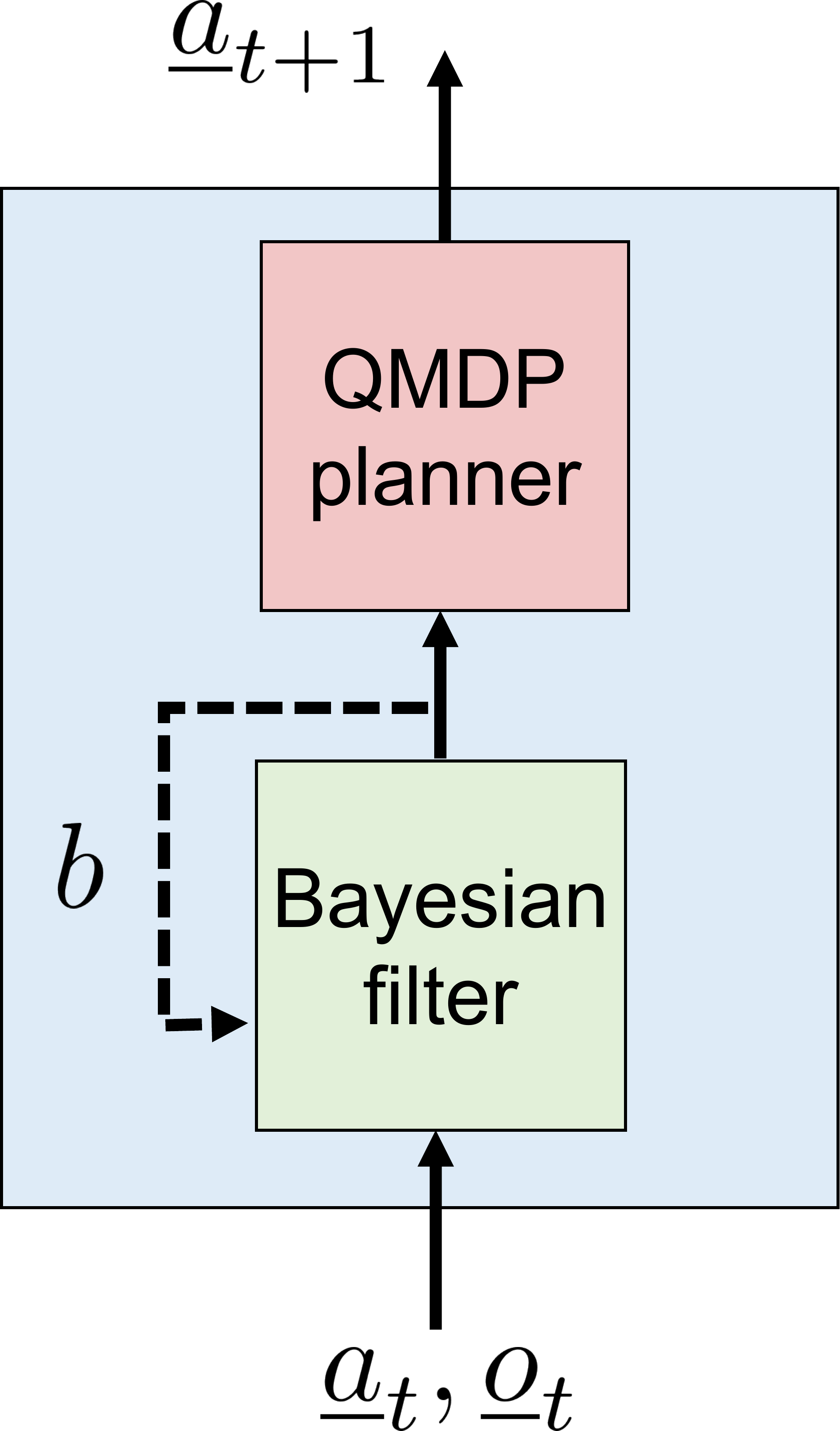}\label{fig:rnn2} }\hspace{3.2mm} 
   \sidesubfloat[][]{ \hspace{-0.27cm}\includegraphics[height=3.0cm]{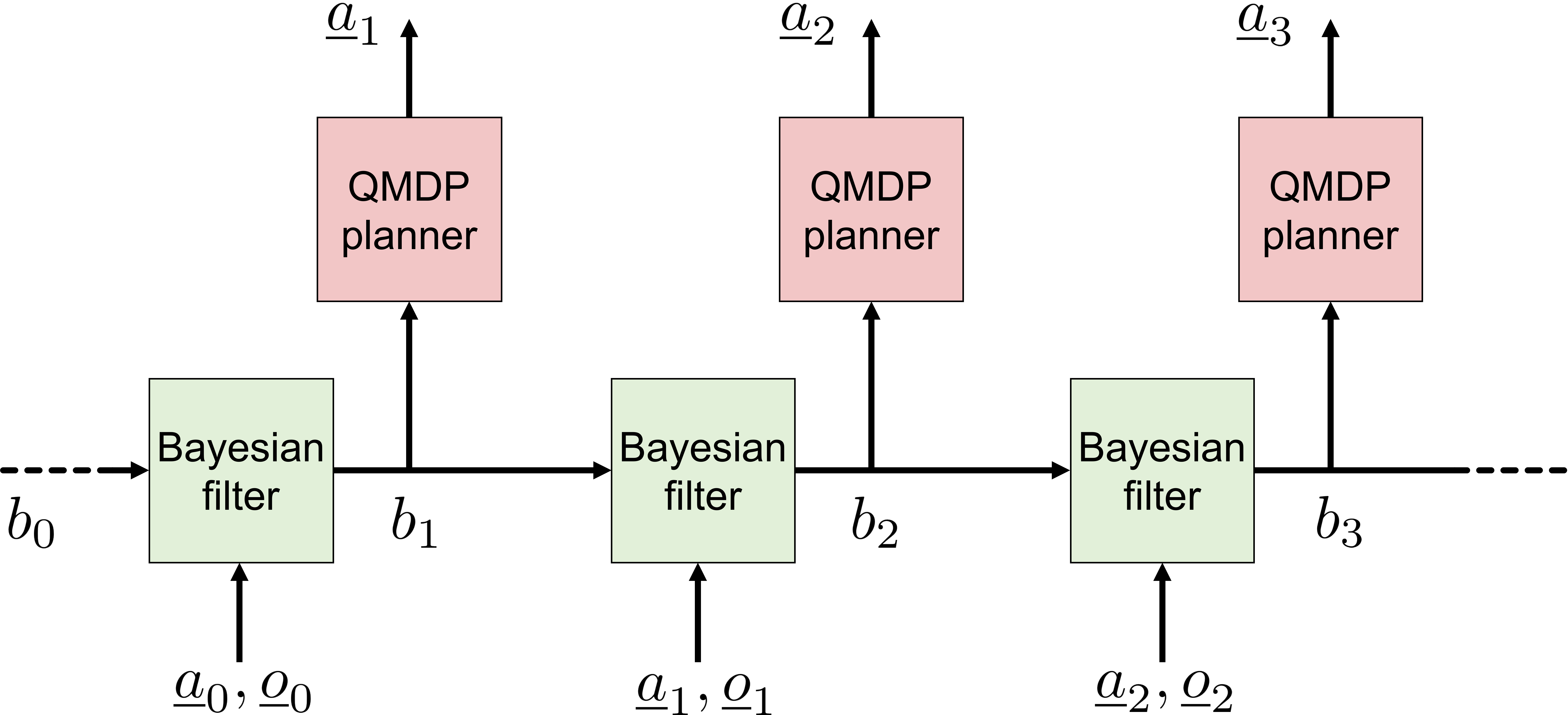}\label{fig:rnn3} } 
   \caption{QMDP-net architecture.  \protect\subref{fig:rnn1} A policy maps a
     history of actions and observations to a new
     action. \protect\subref{fig:rnn2} A QMDP-net is an RNN that imposes
     structure priors for sequential decision making under partial observability. It
     embeds a Bayesian filter and the QMDP algorithm in the network.
     The hidden state of the RNN encodes the belief for POMDP
     planning.  \protect\subref{fig:rnn3} A QMDP-net unfolded in time. }
\label{fig:rnn}
\end{figure*}

A QMDP-net consists of two modules (\figref{fig:rnn}). 
One encodes a Bayesian filter, which performs state estimation by integrating
the past history of agent actions and observations into a belief.  The other
encodes QMDP, a simple, but fast approximate POMDP
planner~\cite{littman1995learning}.  QMDP chooses the agent's actions by
solving the corresponding fully observable Markov decision process (MDP) and
performing one-step look-ahead search on the MDP values weighted by the belief.

\floatsetup[figure]{style=plain,subcapbesideposition=top}

We evaluate the proposed network architecture in an imitation learning
setting. We train on a set of expert trajectories with randomly chosen
task parameter values in \paramset and test with new parameter values.  An expert
trajectory consist of a sequence of demonstrated actions and observations
$(a_1,o_1, a_2, o_2, \ldots )$ for some $\param\in \Theta$. The agent does not
access the ground-truth states or beliefs along the
trajectory during the training.  We define loss as the cross entropy between
predicted and demonstrated action sequences and use RMSProp~\cite{tieleman2012lecture} for training.
See Appendix~\ref{sec:training} for details. Our implementation in Tensorflow~\cite{tensorflow2015-whitepaper}
is available online at {\href{http://github.com/AdaCompNUS/qmdp-net}{http://github.com/AdaCompNUS/qmdp-net}}.

\section{QMDP-Net}\label{sec:qmdpnet}

We assume that all tasks in a parameterized set $\taskset_{\paramset}$ share the
same underlying state space $\taskspace S$, action space $\taskspace A$, and
observation space $\taskspace O$. We want to learn a QMDP-net policy for
$\taskset_{\paramset}$, conditioned on the parameters $\param \in \paramset$.
A QMDP-net is a recurrent policy network. 
The inputs to a QMDP-net are the action $\taskspace a_t\in \taskspace A$ and
the observation $\taskspace o_t\in \taskspace O$ at time step $t$, as well as
the task parameter $\param \in \paramset$.  The output is the action
$\taskspace a_{t+1}$ for time step $t+1$.

A QMDP-net encodes a parameterized POMDP model
$\dense \model(\param) = (S, A, O, T=f_{\sss T}(\cdot | \param), Z=f_{\sss
  Z}(\cdot | \param), R=f_{\sss R}(\cdot | \param))$ and the QMDP algorithm,
which selects actions by solving the model approximately.  We choose
$\modelspace S$, $\modelspace A$, and $\modelspace O$ of $\model(\param)$
manually, based on prior knowledge on $\taskset_\paramset$, specifically, prior knowledge on
$\taskspace S$, $\taskspace A$, and $\taskspace O$.  In general,
$\modelspace S\not= \taskspace S$, $\modelspace A\not= \taskspace A$, and
$\modelspace O\not= \taskspace O$. The model states, actions, and observations
may be abstractions of their real-world counterparts in the task.  In our
robot navigation example (\figref{fig:fig1}), while the robot moves in a
continuous space, we choose $\modelspace S$ to be a grid of finite size. We
can do the same for $\modelspace A$ and $\modelspace O$, in order to reduce
representational and computational complexity.  The transition function $T$,
observation function $Z$, and reward function $R$ of $\model(\param)$ are
conditioned on \param, and are learned from data through end-to-end training.
In this work, we assume that $T$ is the same for all
tasks in $\taskset_\paramset$ to simplify the network architecture.
In other words, $T$ does not depend on \param.

End-to-end training is feasible, because a QMDP-net encodes both a model and
the associated algorithm in a single, fully
differentiable neural network.
The main idea for embedding the algorithm in a neural network is to represent linear
operations, such as matrix multiplication and summation, by convolutional
layers and represent maximum operations by max-pooling layers. Below we
provide some details on the QMDP-net's architecture, which consists of two
modules, a filter and a planner.


\floatsetup[figure]{style=plain, capposition=bottom}
\captionsetup[subfigure]{position=bottom,captionskip=0.3cm}
\begin{figure}[!t]
  \centering
   \subfloat[][{\small Bayesian filter module}]{\hspace{-0.1cm}\includegraphics[height=2.9cm]{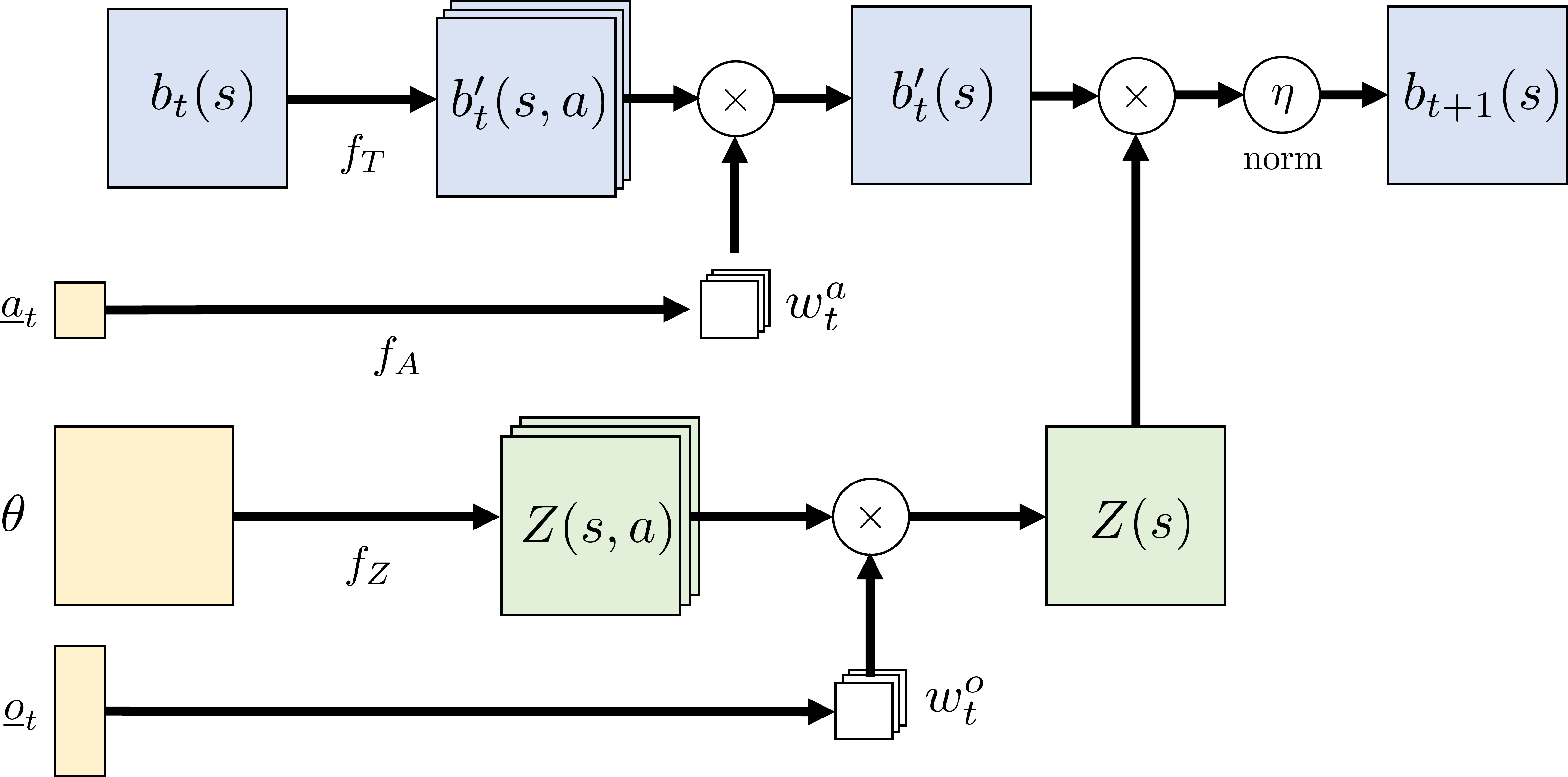}\label{fig:filter}\hspace{0.25cm}}
   \subfloat[][{\small QMDP planner module}]{\hspace{0.25cm} \includegraphics[height=2.9cm]{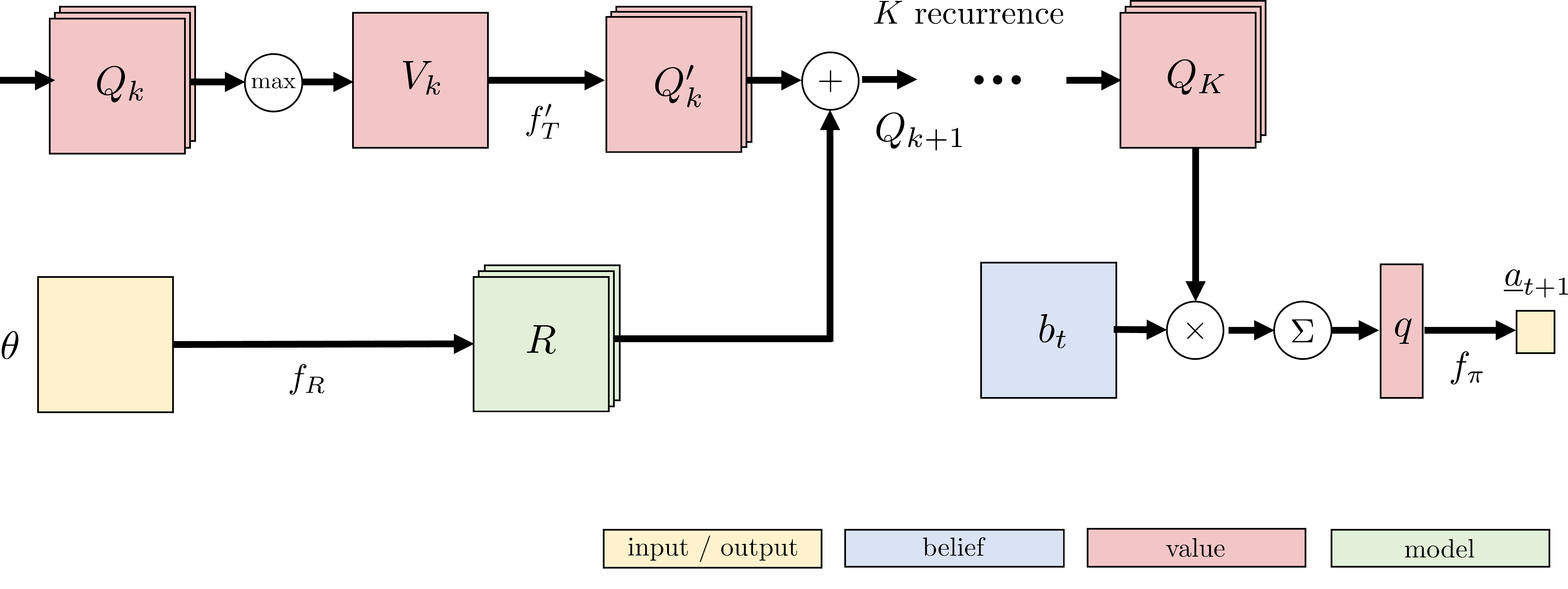}\label{fig:planner}\hspace{-0.1cm} }%
   \caption{A QMDP-net consists of two modules. \protect\subref{fig:filter}
The Bayesian filter module
incorporates the current action $\taskspace{a}_t$ and observation $\taskspace{o}_t$
 into the belief.
\protect\subref{fig:planner} The QMDP planner
module selects the action according to the current belief $b_t$. }
 \label{fig:qmdpnet}
 \end{figure}

\paragraph{Filter module.}
The filter module (\figref{fig:filter}) implements a Bayesian filter. It maps from a belief, action, and observation to a next belief, $b_{t+1} = f(b_{t} | \taskspace{a}_t, \taskspace{o}_t)$. The belief is updated in two steps.
The first accounts for actions, the second for observations:
\begin{equation} \label{eq:filt_T}
\textstyle
b'_t(s) = \sum_{s'\in S}{T(s, \taskspace{a}_{t}, s')b_t(s') },
\end{equation}\vspace{-0.2cm} 
\begin{equation} \label{eq:filt_Z}
\textstyle
b_{t+1}(s) = \eta Z(s, \taskspace{o}_t) b'_t(s),
\end{equation}

where $\taskspace{o}_t \in \taskspace O$ is the observation received after taking action $\taskspace{a}_t \in \taskspace A$ and $\eta$ is a normalization factor.

We implement the Bayesian filter by transforming Eq.~(\ref{eq:filt_T}) and Eq.~(\ref{eq:filt_Z}) to layers of a neural network. 
For ease of discussion consider our $N\mytimes N$ grid navigation task (\figref{fig:fig1}a--c). The agent does not know its own state and only observes neighboring cells. It has access to the task parameter $\param$ that encodes the obstacles, goal, and a belief over initial states. Given the task, we choose $\model(\param)$ to have a $N\mytimes N$ state space. The belief, $b_t(s)$, is now an $N\mytimes N$ tensor. 

Eq. (\ref{eq:filt_T}) is implemented as a convolutional layer with $|A|$ convolutional filters. We denote the convolutional layer by $f_{\sss T}$.  
The kernel weights of $f_{\sss T}$ encode the transition function $T$ in $\model(\param)$.
The output of the convolutional layer, $b_t'(s, a)$, is a $\dense N \mytimes N \mytimes |A|$ tensor. 

$b_t'(s, a)$ encodes the updated belief after taking each of the actions, $a \in A$.
We need to select the belief corresponding to the last action taken by the agent, $\taskspace{a}_t$. 
We can directly index $b_t'(s, a)$ by $\taskspace{a}_t$ if $\modelspace A = \taskspace A$.
In general $\modelspace A \not= \taskspace A$, so we cannot use simple indexing. 
Instead, we will use ``soft indexing''. 
First we encode actions in $\taskspace A$ to actions in $\modelspace A$ through a learned function $f_{\sss A}$.  $f_{\sss A}$ maps from $\taskspace{a}_t$ to an indexing vector $w^{a}_t$, a distribution over actions in $A$. We then weight $b'_t(s, a)$ by $w^{a}_t$ along the appropriate dimension, \ie
\begin{equation}
\textstyle
{b'_t}(s) = \sum_{a \in A}{b'_t(s, a) w^{a}_{t}}.
\end{equation}

Eq. (\ref{eq:filt_Z}) incorporates observations through an observation model $Z(s,o)$. Now $Z(s, o)$ is a $\dense N \mytimes N \mytimes |O|$ tensor that represents the probability of receiving observation $o \in O$ in state $s \in S$. In our grid navigation task observations depend on the obstacle locations.
We condition $Z$ on the task parameter,  $Z(s, o) = f_{\sss Z}(s, o | \param)$ for $\param \in \paramset$.
The function $f_{\sss Z}$ is a neural network, mapping from $\param$ to $Z(s,o)$. In this paper $f_{\sss Z}$ is a CNN. 

$Z(s, o)$ encodes observation probabilities for each of the observations, $o \in O$. 
We need the observation probabilities for the last observation $\taskspace{o}_t$. In general $\modelspace O \not= \taskspace O$ and we cannot index $Z(s, o)$ directly. Instead, we will use soft indexing again. We encode observations in $\taskspace O$ to observations in $\modelspace O$ through~$f_{\sss O}$.  $f_{\sss O}$ is a function mapping from $\taskspace{o}_t$ to an indexing vector, $w^{o}_t$, a distribution over $O$. We then weight $Z(s, o)$ by $w^{o}_t$, \ie
\begin{equation}
\textstyle
Z(s) = \sum_{o \in O}Z(s,o)w^{o}_t.
\end{equation}

Finally, we obtain the updated belief, $b_{t+1}(s)$,  by multiplying ${b'_t}(s)$ and $Z(s)$ element-wise, and normalizing over states.
In our setting the initial belief for the task $\task(\param)$ is encoded in $\param$. We initialize the belief in QMDP-net through an additional encoding function, $b_0 = f_{\sss B}(\param)$.  

\paragraph{Planner module.} 
The QMDP planner~(\figref{fig:planner}) performs value iteration at its core. $Q$ values are computed by iteratively applying Bellman updates,
\begin{equation} \label{eq:Q}
\textstyle
Q_{k+1}(s, a) =  R(s,a) + \gamma\sum_{s' \in S}{T(s, a, s') V_k(s')},
\end{equation}\vspace{-0.25cm}
\begin{equation} \label{eq:V}
\textstyle
V_k(s) = \max_a{Q_k(s,a)}.
\end{equation}\vspace{-0.4cm}

Actions are then selected by weighting the $Q$ values with the belief.

We can implement value iteration using convolutional and max pooling layers~\cite{tamar2016value, shankar2016reinforcement}. In our grid navigation task $Q(s,a)$ is a $N\mytimes N \mytimes |A|$ tensor. 
Eq.~(\ref{eq:V}) is expressed by a max pooling layer, where $Q_k(s, a)$ is the input and $V_k(s)$ is the output. 
Eq.~(\ref{eq:Q}) is a $N\mytimes N$ convolution with $|A|$ convolutional filters, followed by an addition operation with $R(s,a)$, the reward tensor. We denote the convolutional layer by $f_{\sss T}'$. The kernel weights of $f_{\sss T}'$ encode the transition function $T$, similarly to $f_{\sss T}$ in the filter. Rewards for a navigation task depend on the goal and obstacles. We condition rewards on the task parameter, $R(s, a) = f_{\sss R}(s, a | \param)$. $f_{\sss R}$ maps from $\param$ to $R(s,a)$. In this paper $f_{\sss R}$ is a CNN.

We implement $K$ iterations of Bellman updates by stacking the layers representing Eq.~(\ref{eq:Q}) and Eq.~(\ref{eq:V}) $K$ times with tied weights. After $K$ iterations we get $Q_K(s,a)$, the approximate $Q$ values for each state-action pair. We weight the $Q$ values by the belief to obtain action values,
\begin{equation} 
\textstyle 
q(a) = \sum_{s \in S}{Q_K(s, a) b_t(s)}.
\end{equation}
Finally, we choose the output action through a low-level policy function, $f_{\sss \pi}$, mapping from $q(a)$ to the action output, $\taskspace{a}_{t+1}$.

QMDP-net naturally extends to higher dimensional discrete state
 spaces (\eg{} our maze navigation task) where $n$-dimensional convolutions 
 can be used~\cite{ji2013}. While $\model(\param)$ is restricted to a discrete space, 
 we can handle continuous tasks $\taskset_{\paramset}$ by simultaneously learning a discrete $\model(\param)$ for planning, 
 and $f_{\sss A}, f_{\sss O}, f_{\sss B}, f_{\sss \pi}$ to map between states, actions and observations in $\taskset_{\paramset}$ and $\model(\param)$.


\section{Experiments}\label{sec:experiments}

The main objective of the experiments is to understand the benefits of structure
priors on learning neural-network policies.  We create several alternative
network architectures by gradually relaxing the structure priors and evaluate
the architectures on simulated robot navigation and manipulation tasks.
While these tasks are simpler than, for example, Atari games, in terms of
visual perception, they are in fact very challenging, because of the
sophisticated long-term reasoning required to handle partial observability and
distant future rewards.  Since the exact state of the robot is unknown, a
successful policy must reason over many steps to gather information and
improve state estimation through partial and noisy observations.  It also must
reason about the trade-off between the cost of information gathering and the
reward in the distance future.

\subsection{Experimental Setup}\label{sec:domains}
We compare the QMDP-net with a number of related alternative architectures. 
Two are QMDP-net variants.  \emph{Untied QMDP-net} relaxes the constraints on
the planning module by untying the weights representing the state-transition
function over the different CNN layers.  \emph{LSTM QMDP-net} replaces the
filter module with a generic LSTM module.  The other two architectures do not
embed POMDP structure priors at all.  \emph{CNN+LSTM} is a state-of-the-art
deep CNN connected to an LSTM. It is similar to the DRQN
architecture proposed for reinforcement learning
under partially observability~\cite{hausknecht2015deep}.  \emph{RNN} is a
basic recurrent neural network with a single fully-connected hidden layer.
RNN contains no structure specific to planning under partial
observability.

Each experimental domain contains a parameterized set of tasks
$\taskset_{\paramset}$. The parameters \param encode an environment, a goal,
and a belief over the robot's initial state. To train a policy for
$\taskset_{\paramset}$, we generate random environments, goals, and initial
beliefs.
We construct ground-truth POMDP models for the
generated data and apply the QMDP algorithm.  If the QMDP algorithm
successfully reaches the goal, we then retain the resulting sequence of action
and observations $(a_1, o_1, a_2, o_2, \ldots )$ as an expert trajectory,
together with the corresponding environment, goal, and initial belief.  It is
important to note that  the ground-truth POMDPs are used only for generating
expert trajectories and not for learning the QMDP-net.

For fair comparison, we train all networks
using the same set of expert trajectories in each domain.
We perform basic search over training parameters, the number
of layers, and the number of hidden units for each network architecture.
Below we briefly describe the experimental domains.
See Appendix~\ref{sec:app_details} for implementation  details.

\vspace*{-6pt}
\paragraph{Grid-world navigation.}
A robot navigates in an unknown building given a floor map and a goal. The
robot is uncertain of its own location. It is equipped with a LIDAR that
detects obstacles in its direct neighborhood. The world is uncertain:  the 
robot may fail to execute desired actions, possibly because of wheel slippage,
and  the LIDAR may produce false readings.
We implemented a simplified version of this task in a discrete $n\mytimes n$
grid world (\figref{fig:fig1}c).  The task parameter $\param$ is represented
as an $n\mytimes n$ image with three channels.  The first channel encodes the
obstacles in the environment, the second channel encodes the goal, and the
last channel encodes the belief over the robot's initial state.  The robot's
state represents its position in the grid. It has five actions:
moving in each of the four canonical directions or staying put. The LIDAR
observations are compressed into four binary values corresponding to obstacles
in the four neighboring cells.  We consider both a deterministic and a
stochastic variant of the domain.  The stochastic variant adds action and
observation uncertainties.  The robot fails to execute the specified move
action and stays in place with probability $0.2$.  The observations are faulty
with probability $0.1$ independently in each direction. We trained a policy
using expert trajectories from $10,000$ random environments, $5$
trajectories from each environment.  We then tested on a separate set of $500$
random environments.

\vspace*{-6pt}
\paragraph{Maze navigation.}
\begin{wrapfigure}[14]{r}{5.8cm}  
  \centering
  \vspace{-10pt}
  \includegraphics[width=2.6cm]{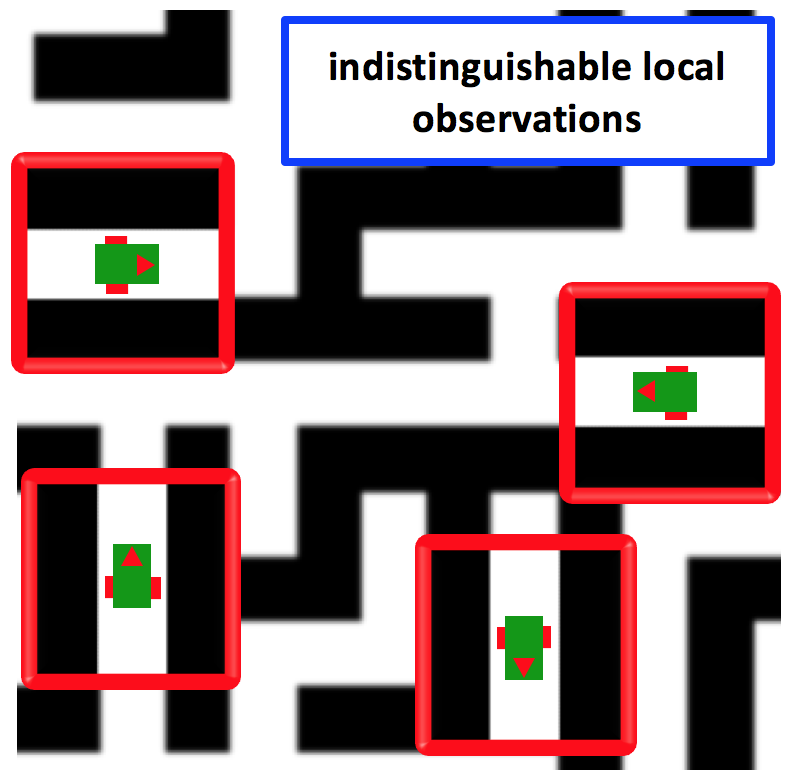}
  \caption{
    Highly ambiguous observations in a maze. The four observations (in red)
    are the same, despite that the robot states are all different.
    \vspace{-0.3cm}}
  \label{fig:maze} 
\end{wrapfigure}
A differential-drive robot navigates in a maze with the help of a map, but it
does not know its pose (\figref{fig:fig1d}).  This domain is similar to the
grid-world navigation, but it is significant more challenging.  The robot's
state contains both its position and orientation.  The robot cannot move
freely because of kinematic constraints. It has four actions: move forward,
turn left, turn right and stay put.  The observations are relative to the
robot's current orientation,
and the increased ambiguity makes it more difficult to
localize the robot, especially when the initial state is highly uncertain.
Finally, successful trajectories in mazes are typically much longer than those
in randomly-generated grid worlds.
Again we trained on expert trajectories in $10,000$ randomly
generated mazes and tested them in $500$ new ones.

\vspace*{-6pt}
\paragraph{2-D object grasping.}

\floatsetup[figure]{style=plain,subcapbesideposition=top}
\begin{wrapfigure}[10]{r}{5.8cm}
  \vspace*{-36pt}  
  \centering
  \sidesubfloat[][]{\hspace{-0.0cm}\includegraphics[height=2.5cm]{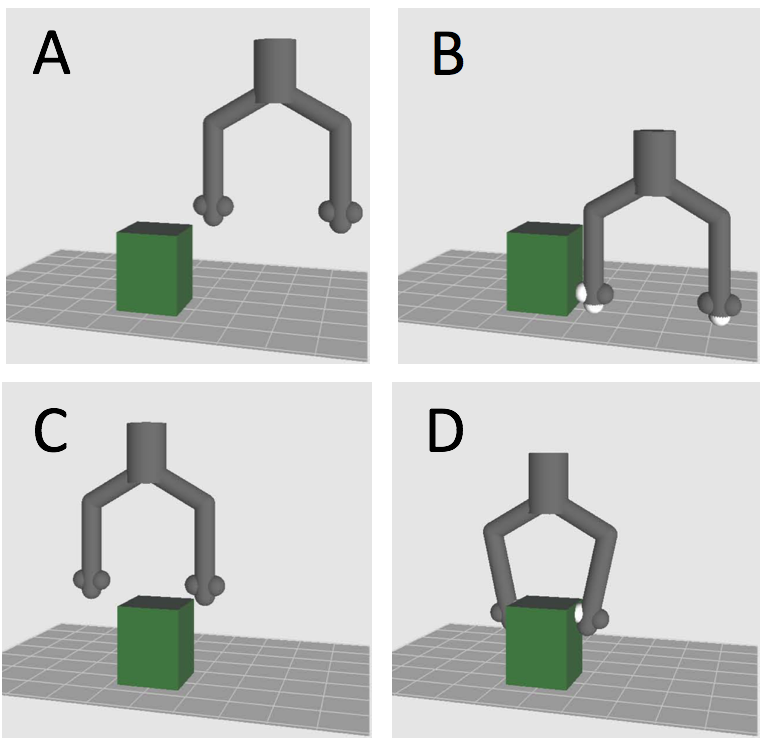}\label{fig:grasp_a}\hspace{0.2cm}} 
  \sidesubfloat[][]{\hspace{-0.0cm}\includegraphics[height=2.5cm]{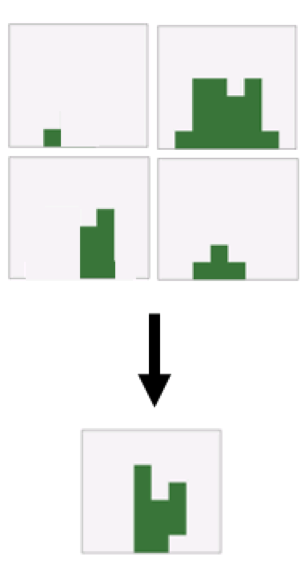}\label{fig:grasp_b}} 
  \caption{Object grasping using touch sensing. \protect\subref{fig:grasp_a}
    An example \cite{bai2010monte}. \protect\subref{fig:grasp_b} Simplified
    2-D object grasping. Objects from
    the training set (top) and the test set (bottom). }
\label{fig:grasp}
\end{wrapfigure}

A robot gripper picks up novel objects from a table using a two-finger hand
with noisy touch sensors at the finger tips.  The gripper
uses the fingers to perform
compliant motions while maintaining contact with the object or to grasp the
object.  It knows the shape of the object to be grasped, maybe
from an object database. However, it does not know its own pose relative to
the object and relies on the touch sensors to localize itself.
We implemented a simplified 2-D variant of this task,  modeled as a
POMDP~\cite{hsiao2007grasping}.
The task parameter \param is an image with three channels encoding
the object shape, the grasp point, and a belief over the gripper's initial
pose.  The gripper has four actions, each moving in a  canonical direction
unless it touches the object or the environment boundary. 
Each finger has $3$ binary touch sensors at the tip, resulting in
$64$ distinct observations. 
We trained on expert demonstration on $20$ different objects with $500$ randomly
sampled poses for each object. We then tested on $10$ previously unseen objects
in random poses.


\subsection{Choosing QMDP-Net Components for a Task} 
Given a new task $\taskset_{\paramset}$, we need to choose an appropriate neural network representation for $\model(\param)$. More specifically, we need to choose $S, A$ and $O$, and a representation for the functions $f_{\sss R}, f_{\sss T}, f_{\sss T}', f_{\sss Z}, f_{\sss O}, f_{\sss A}, f_{\sss B}, f_{\sss \pi}$. This provides an opportunity to incorporate domain knowledge in a principled way. For example, if $\taskset_{\paramset}$ has a local and spatially invariant connectivity structure, we can choose convolutions with small kernels to represent $f_{\sss T}$, $f_{\sss R}$ and $f_{\sss Z}$.

In our experiments we use $\dense S=N\mytimes N$ for $\dense N\mytimes N$ grid navigation, and $\dense S=N\mytimes N\mytimes 4$ for $\dense N\mytimes N$ maze navigation where the robot has $4$ possible orientations. We use $|\modelspace A| = |\taskspace A|$ and $|\modelspace O| = |\taskspace O|$ for all tasks except for the object grasping task, where $|\taskspace O| = 64$ and $|\modelspace O|=16$.
We represent $f_{\sss T}, f_{\sss R}$ and $f_{\sss Z}$ by CNN components with $3\mytimes 3$ and $5\mytimes 5$ kernels depending on the task. We enforce that $f_{\sss T}$ and $f_{\sss Z}$ are proper probability distributions by using softmax and sigmoid activations on the convolutional kernels, respectively. Finally, $f_{\sss O}$ is a small fully connected component, $f_{\sss A}$ is a one-hot encoding function, $f_{\sss \pi}$ is a single softmax layer, and $f_{\sss B}$ is the identity function. 

We can adjust the \emph{amount of planning} in a QMDP-net by setting $K$. A large $K$ allows propagating information to more distant states without affecting the number of parameters to learn. However, it results in deeper networks that are computationally expensive to evaluate and more difficult to train.
We used $K=20\dots116$ depending on the problem size. We were able to transfer policies to larger environments by increasing $K$ up to $450$ when executing the policy.

In our experiments the representation of the task parameter $\param$ is isomorphic to the chosen state space~$S$. 
While the architecture is not restricted to this setting, we rely on it to represent $\dense f_{\sss T}, f_{\sss Z}, f_{\sss R}$ by convolutions with small kernels. Experiments with a more general class of problems is an interesting direction for future work.

\subsection{Results and Discussion}\label{sec:discussion}
The main results are reported in \tabref{tab:all}. Some additional results are
reported in Appendix~\ref{sec:app_results}.  For each domain, we report the
task success rate and the average number of time steps for task completion.
Comparing the completion time is meaningful only when the success rates are
similar.

\paragraph{QMDP-net successfully learns policies that generalize to new environments.}  
When evaluated on new environments, the QMDP-net has higher success rate
and faster completion time than the alternatives in nearly all domains.
To understand better the performance difference, we specifically compared the
architectures in a fixed environment for navigation.  Here only the initial
state and the goal vary across the task instances, while the environment
remains the same. See the results in the
last row of \tabref{tab:all}.  The QMDP-net and the alternatives have
comparable performance. Even RNN performs very well. Why? In a fixed
environment, a network may learn the features of an optimal policy directly,
\eg, going straight towards the goal.  In contrast, the QMDP-net learns a model
for \emph{planning}, \ie, generating a near-optimal policy for a given
arbitrary environment.

\paragraph{POMDP   structure  priors improve the  performance of  learning complex policies.}  
Moving across \tabref{tab:all} from left to right, we gradually relax the
POMDP structure priors on the network architecture.
As the structure priors weaken, so does the overall performance.  However,
strong priors sometimes over-constrain the network and result in degraded
performance.  For example, we found that tying the weights of
$f_{\sss T}$ in the filter and $f_{\sss T}'$ in the
planner may lead to worse policies. While both $f_{\sss T}$ and $f_{\sss T}'$ represent the same
underlying transition dynamics, using different weights allows each to choose
its own approximation and thus greater flexibility.
We shed some light on this issue and visualize the learned POMDP model in Appendix~\ref{sec:app_visualize}.

\floatsetup[table]{capposition=top}
\begin{table}[!t]
  \caption{Performance comparison of QMDP-net and alternative
    architectures for recurrent policy networks.  SR is
    the success rate in percentage. Time is
    the average number of time steps for task completion.
    D-$n$ and S-$n$ denote deterministic and stochastic variants of a domain
    with environment size $n\mytimes n$.
  }
\label{tab:all}
\begin{center}
 \begin{footnotesize}
\resizebox{0.94\textwidth}{!}{%
  \begin{tabular}{@{\hskip 4pt}lLRLRLRLRLRLR@{\hskip 4pt}}
\toprule
  \multicolumn{1}{c}{}  & \multicolumn{2}{c}{\textbf{QMDP}}  &  \multicolumn{2}{c}{\textbf{QMDP-net} }   &  \multicolumn{2}{c}{\textbf{Untied}}
  &  \multicolumn{2}{c}{\textbf{LSTM}} &  \multicolumn{2}{c}{\textbf{CNN}}    &  \multicolumn{2}{c}{\textbf{RNN}}  
  \\
  & \multicolumn{2}{c}{}  &  \multicolumn{2}{c}{}   &  \multicolumn{2}{c}{\textbf{QMDP-net}}
  &  \multicolumn{2}{c}{\textbf{QMDP-net}} &  \multicolumn{2}{c}{\textbf{+LSTM}}    &  \multicolumn{2}{c}{}  
  \\
  \abovespace
 \textbf{Domain}    & SR  & {Time}  & SR & {Time}   & SR & {Time}   & SR & {Time}   & SR & {Time}   & SR & {Time}  \\
\midrule 
Grid D-10 
   & 99.8  & 8.8        & 99.6  & 8.2       & 98.6  & 8.3        & 84.4  & 12.8        &  90.0 & 13.4     & 87.8  & 13.4   \\  
Grid D-18 
   &  99.0 & 15.5      & 99.0   & 14.6    & 98.8  & 14.8       & 43.8  & 27.9        &  57.8  & 33.7   & 35.8  & 24.5      \\  
Grid D-30 
    & 97.6  & 24.6    & 98.6  & 25.0      & 98.8  & 23.9     & 22.2  & 51.1        &  19.4  &  45.2    & 16.4  & 39.3   \\  
\abovespace
Grid S-18 
    & 98.1  & 23.9     & 98.8 &  23.9    & 95.9  & 24.0       & 23.8  & 55.6        & 41.4 & 65.9    & 34.0  & 64.1   \\  
\abovespace
Maze D-29 
    &  63.2  & 54.1          & 98.0   &56.5      & 95.4  & 62.5      & 9.8   &  57.2        & 9.2   & 41.4       &  9.8  &  47.0      \\  
Maze S-19 
    & 63.1  & 50.5       & 93.9 & 60.4     & 98.7  & 57.1       & 18.9  & 79.0           & 19.2  &  80.8        & 19.6 & 82.1   \\  
\abovespace
Hallway2
    & 37.3   & 28.2      & 82.9    &  64.4     &  69.6   &  104.4       & 82.8  & 89.7        & 77.8 & 99.5      & 68.0 & 108.8    \\  
\abovespace
Grasp
    &   98.3   & 14.6   & 99.6  &  18.2     & 98.9  & 20.4       & 91.4  & 26.4         & 92.8 & 22.1     & 94.1 & 25.7   \\  
\abovespace
Intel Lab
  & 90.2  & 85.4        & 94.4  & 107.7       &  20.0  & 55.3     & \multicolumn{2}{c}{-}        & \multicolumn{2}{c}{-}       & \multicolumn{2}{c}{-}   \\ 
Freiburg
    & 88.4  & 66.9    & 93.2  & 81.1       &  37.4  & 51.7     & \multicolumn{2}{c}{-}        & \multicolumn{2}{c}{-}       & \multicolumn{2}{c}{-}   \\ 
  \abovespace
Fixed grid
  &  98.8 & 17.4      &  98.6   & 17.6          &99.8 & 17.0       & 97.0  & 19.7       &  98.4  & 19.9     & 98.0 & 19.8   \\  
\bottomrule 
\end{tabular}
}
\end{footnotesize}
\end{center}
\end{table}

\paragraph{QMDP-net learns ``incorrect'', but useful  models.}
Planning under partial observability is intractable in general, and we must
rely on approximation algorithms.  A QMDP-net encodes both a POMDP model and
QMDP, an approximate POMDP algorithm that solves the model. We then train the
network end-to-end. This provides the opportunity to learn an ``incorrect'',
but useful model that compensates the limitation of the approximation
algorithm, in a way similar to reward shaping in
reinforcement learning~\cite{ng1999policy}.  Indeed, our results show that the QMDP-net achieves
higher success rate than QMDP in nearly all tasks. In particular, QMDP-net
performs well on the well-known Hallway2 domain, which is designed to expose
the weakness of QMDP resulting from its myopic planning horizon.  The planning
algorithm is the same for both the QMDP-net and QMDP, but the QMDP-net learns a more
effective model from expert demonstrations.  This is true even though QMDP
generates the expert data for training.  We note that the expert data contain
only successful QMDP demonstrations.  When both successful and unsuccessful
QMDP demonstrations were used for training, the QMDP-net did not perform
better than QMDP, as one would expect.

\paragraph{QMDP-net policies learned in small environments
  transfer directly to larger environments.}
Learning a policy for large  environments from scratch is often difficult.
A more scalable approach would be to learn a policy in small environments and
transfer it to large environments by repeating the reasoning process.
To transfer a learned QMDP-net policy, we simply expand its planning module by
adding more recurrent layers.
Specifically, we trained a policy in randomly generated $\dense 30\times30$
grid worlds with $K=90$. We then set $K=450$ and applied the learned policy to
several real-life environments,
including Intel Lab ($100\mytimes 101$)  and Freiburg ($139\mytimes 57$),   
using their  LIDAR maps (\figref{fig:fig1}c) from
the Robotics Data Set Repository
\cite{radish}. See the results for these two environments in
\tabref{tab:all}. Additional results with different $K$
settings and other buildings are available in Appendix~\ref{sec:app_results}.

\section{Conclusion}\label{sec:conclusion}
A QMDP-net is a deep recurrent policy network that embeds POMDP
structure priors for planning under partial observability.  While generic
neural networks learn a direct mapping from inputs to outputs, QMDP-net learns
how to \emph{model} and \emph{solve} a planning task.  The network is fully
differentiable and allows for end-to-end training.

Experiments on several simulated robotic tasks show that learned QMDP-net
policies successfully generalize to new environments and transfer to larger
environments as well.  The POMDP structure priors and end-to-end training
substantially improve the performance of learned policies.  Interestingly,
while a QMDP-net encodes the QMDP algorithm for
planning, learned QMDP-net policies sometimes outperform QMDP. 

There are many exciting directions for future exploration.
First, a major limitation of our current approach is the state space
representation. 
The value iteration algorithm used in QMDP iterates through the entire state
space and is well known to suffer from  the ``curse of dimensionality''.
To alleviate this difficulty,  the QMDP-net, through end-to-end training,
may learn a
much smaller  abstract state space representation for planning. 
One may also incorporate hierarchical planning~\cite{gupta2017cognitive}. 
Second,
QMDP  makes strong approximations in order to reduce computational
complexity. We want to explore the possibility of  embedding
more sophisticated POMDP algorithms in the network architecture.
While these algorithms provide stronger planning
performance, their algorithmic sophistication increases the difficulty of
learning. 
Finally,  we have so far restricted the work to imitation learning. It would be
exciting to extend it to reinforcement learning. Based on
earlier work~\cite{shankar2016reinforcement,tamar2016value},  this is  indeed
promising.


\medskip
\begin{small}
\paragraph{Acknowledgments}
We thank Leslie Kaelbling and Tom\'as Lozano-P\'erez for insightful
discussions that helped to improve our understanding of the problem. 
The work is supported in part by Singapore Ministry of
  Education AcRF grant MOE2016-T2-2-068 and National University of
  Singapore AcRF grant R-252-000-587-112.
  

\bibliography{manual}
\bibliographystyle{abbrvnat}

\end{small}

\clearpage
\appendix


\section{Supplementary Experiments}\label{sec:app_results}

\subsection{Navigation on Large LIDAR Maps}

We provide results on additional environments for the LIDAR map navigation task. LIDAR maps are obtained from \cite{unibonn}. See \secref{sec:app_real} for details. \textbf{Intel} corresponds to Intel Research Lab. \textbf{Freiburg} corresponds to Freiburg, Building 079. \textbf{Belgioioso} corresponds to Belgioioso Castle. \textbf{MIT} corresponds to the western wing of the MIT CSAIL building. We note the size of the grid size $NxM$ for each environment.  
A QMDP-net policy is trained on the $30x30$-D grid navigation domain on randomly generated environments using $K=90$. 
We then execute the learned QMDP-net policy with different $K$ settings, \ie we add convolutional layers to the planner that share the same kernel weights. We report the
task success rate and the average number of time steps for task completion.

\floatsetup[table]{capposition=top}
\begin{table}[H]
\caption{Additional results for navigation on large LIDAR maps.}
\label{tab:app_buildings}
\begin{center}
\begin{footnotesize}
\begin{tabular}{@{\hskip 4pt}lLRLRLRLRLR@{\hskip 4pt}}
\toprule
  & \multicolumn{2}{c}{\textbf{QMDP}}  &  \multicolumn{2}{c}{\textbf{QMDP-net}}   &  \multicolumn{2}{c}{\textbf{QMDP-net}}
  &  \multicolumn{2}{c}{\textbf{QMDP-net}} &  \multicolumn{2}{c}{\textbf{Untied}} 
  \\
  & \multicolumn{2}{c}{}  						&  \multicolumn{2}{c}{K=450} 			&  \multicolumn{2}{c}{K=180}
  &  \multicolumn{2}{c}{K=90} 		&  \multicolumn{2}{c}{\textbf{QMDP-net}}
  \\
\textbf{Domain}        & SR  & {Time}   & SR & {Time}    & SR & {Time}    & SR & {Time}    & SR & {Time} \\
\midrule
Intel $100\mytimes101$
   & 90.2  & 85.4        & \textbf{94.4}  & 108.0       & 83.4  & 89.6           & 40.8  & 78.6          & 20.0  & 55.3     \\
Freiburg $139\mytimes57$
    & 88.4  & 66.9    & 92.0  & 91.4                    & \textbf{93.2}  & 81.1    & 55.8  & 68.0       & 37.4  & 51.7      \\
Belgioioso $151\mytimes35$
    & 95.8  & 63.9        & \textbf{95.4}  & 71.8      & 90.6  & 62.0             & 60.0 & 54.3        & 41.0  & 47.7      \\
MIT $41\mytimes83$
   &  94.4 & 42.6     & 91.4  & 53.8               & \textbf{96.0}  & 48.5       & 86.2 & 45.4       & 66.6  & 41.4       \\
\bottomrule
\end{tabular}
\end{footnotesize}
\end{center}
\end{table}

 In the conventional setting, when value iteration is executed on a fully known MDP, increasing $K$ improves the value function approximation and improves the policy in return for the increased computation. In a QMDP-net increasing $K$ has two effects on the overall planning quality. Estimation accuracy of the latent values increases and reward information can propagate to more distant states. On the other hand the learned latent model does not necessarily fit the true underlying model, and it can be overfitted to the $K$ setting during training. Therefore a too high $K$ can degrade the overall performance. We found that  $K_{test} = 2K_{train}$ significantly improved success  rates in all our test cases. Further increasing $K_{test} = 5K_{train}$ was beneficial in the Intel and Belgioioso environments, but it slightly decreased success rates for the Freiburg and MIT environments.

We compare QMDP-net to its untied variant, Untied QMDP-net. We cannot expand the layers of Untied QMDP-net during execution. In consequence, the performance is poor. Note that the other alternative architectures we considered are specific to the input size and thus they are not applicable.

\subsection{Learning ``Incorrect'' but Useful Models}

We demonstrate that an ``incorrect'' model can result in better policies when solved by the approximate QMDP algorithm. We compute QMDP policies on a POMDP with modified reward values, then evaluate the policies using the original rewards. We use the deterministic $29\mytimes 29$ maze navigation task where QMDP did poorly. We attempt to shape rewards manually. Our motivation is to break symmetry in the model, and to implicitly encourage information gathering and compensate for the one-step look-ahead approximation in QMDP.  \textbf{Modified 1.} We increase the cost for the stay actions to $20$ times of its original value. \textbf{Modified 2.} We increase the cost for the stay action to $50$ times of its original value, and the cost for the turn right action to $10$ times of its original value. 

\floatsetup[table]{capposition=top}
\setlength\tabcolsep{2mm}
\begin{table}[H]
\caption{QMDP policies computed on  an ``incorrect'' model and evaluated on the ``correct'' model.}\label{tab:rewardshaping}
\begin{center}
\begin{footnotesize}
\begin{tabular}{lccc}
\toprule
  &  &  & Original
  \\
\textbf{Variant}        & SR  & Time  & reward  \\
\midrule
Original
      &  63.2 &  54.1  &  1.09  \\
Modified 1
      &  65.0  & 58.1    &  1.71\\
Modified 2
      & \textbf{ 93.0}  & 71.4    &  4.96  \\
\bottomrule
\end{tabular}
\end{footnotesize}
\end{center}
\vskip -0.1in
\end{table}

Why does the ``correct'' model result in poor policies when solved by QMDP? At a given point the $Q$ value for a set of possible states may be high for the turn left action and low for the turn right action; while for another set of states it may be the opposite way around. In expectation, both next states have lower value than the current one, thus the policy chooses the stay action, the robot does not gather information and it is stuck in one place. Results demonstrate that planning on an ``incorrect'' model may improve the performance on the ``correct'' model.

\section{Visualizing the Learned Model}\label{sec:app_visualize}
\subsection{Value Function} 

We plot the value function predicted by a QMDP-net for the  $\dense 18 \times 18$ stochastic grid navigation task. We used $K=54$ iterations in the QMDP-net. As one would expect, states close to the goal have high values.

\floatsetup[figure]{style=plain,subcapbesideposition=top}
\begin{figure}[H]
  \centering
	\includegraphics[width=0.6\textwidth]{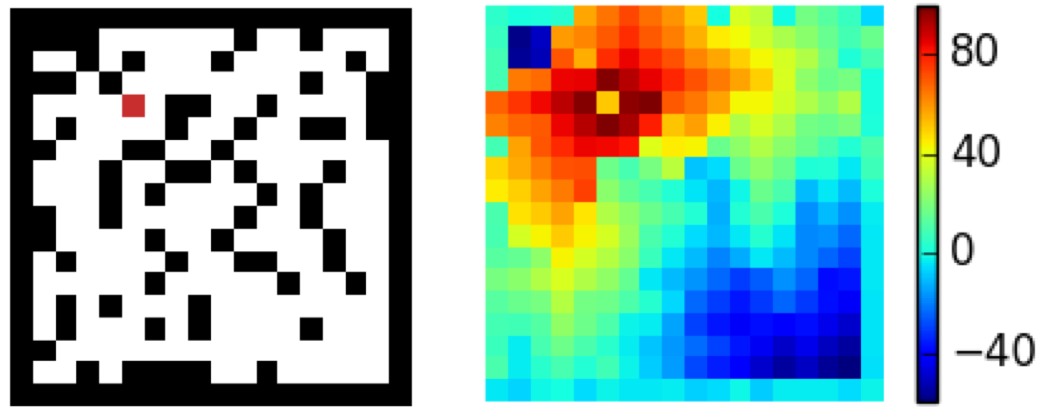}
  \caption{ Map of a test environment and the corresponding learned value function $V_K$.
}
\label{fig:app_V}
\end{figure}

\subsection{Belief Propagation} 

\begin{figure}[H]
  \centering
	\includegraphics[width=0.98\textwidth]{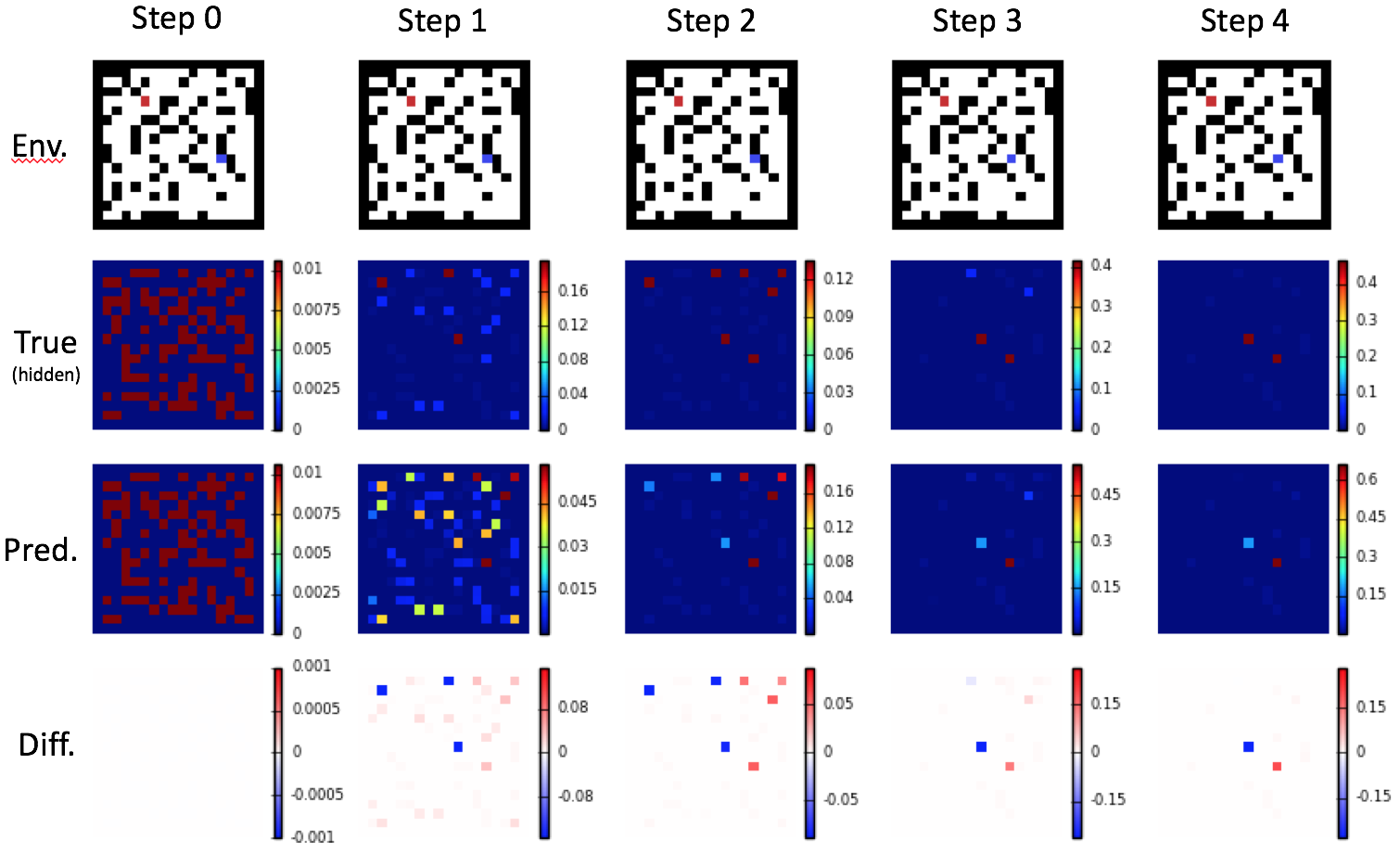}
  \caption{Policy execution and belief propagation in the  $\dense 18 \times 18$ stochastic grid navigation task. }
\label{fig:app_beliefs}
\end{figure}

We plot the execution of a learned QMDP-net policy and the internal belief propagation on the  $\dense 18 \times 18$ stochastic grid navigation task.  The first row in \figref{fig:app_beliefs} shows the environment including the goal (red) and the unobserved pose of the robot (blue). The second row shows ground-truth beliefs for reference. We do not access ground-truth beliefs during training except for the initial belief. The third row shows beliefs predicted by a QMDP-net. The last row shows the difference between the ground-truth and predicted beliefs.

\floatsetup[figure]{style=plain,subcapbesideposition=top}

The figure demonstrates that QMDP-net was able to learn a reasonable filter for state estimation in a noisy environment. In the depicted example the initial belief is uniform over approximately half of the state space (Step 0). Due to the highly uncertain initial belief and the observation noise the robot stays in place for two steps (Step 1 and 2). After two steps the state estimation is still highly uncertain, but it is mostly spread out right from the goal. Therefore, moving 
left is a reasonable choice (Step 3). After an additional stay action (Step 4) the belief distribution is small enough and the robot starts moving towards the goal (not shown).

\subsection{State-Transition Function} 
We plot the learned and ground-truth state-transition functions. Columns of the table correspond to actions. The first row shows the ground-truth transition function. The second row shows $f_{\sss T}$, the learned state-transition function in the filter. The third row shows $f_{\sss T}'$, the learned state-transition function in the planner. 

\floatsetup[figure]{style=plain,subcapbesideposition=top}
\begin{figure}[H]
  \centering
	\includegraphics[width=0.7\textwidth]{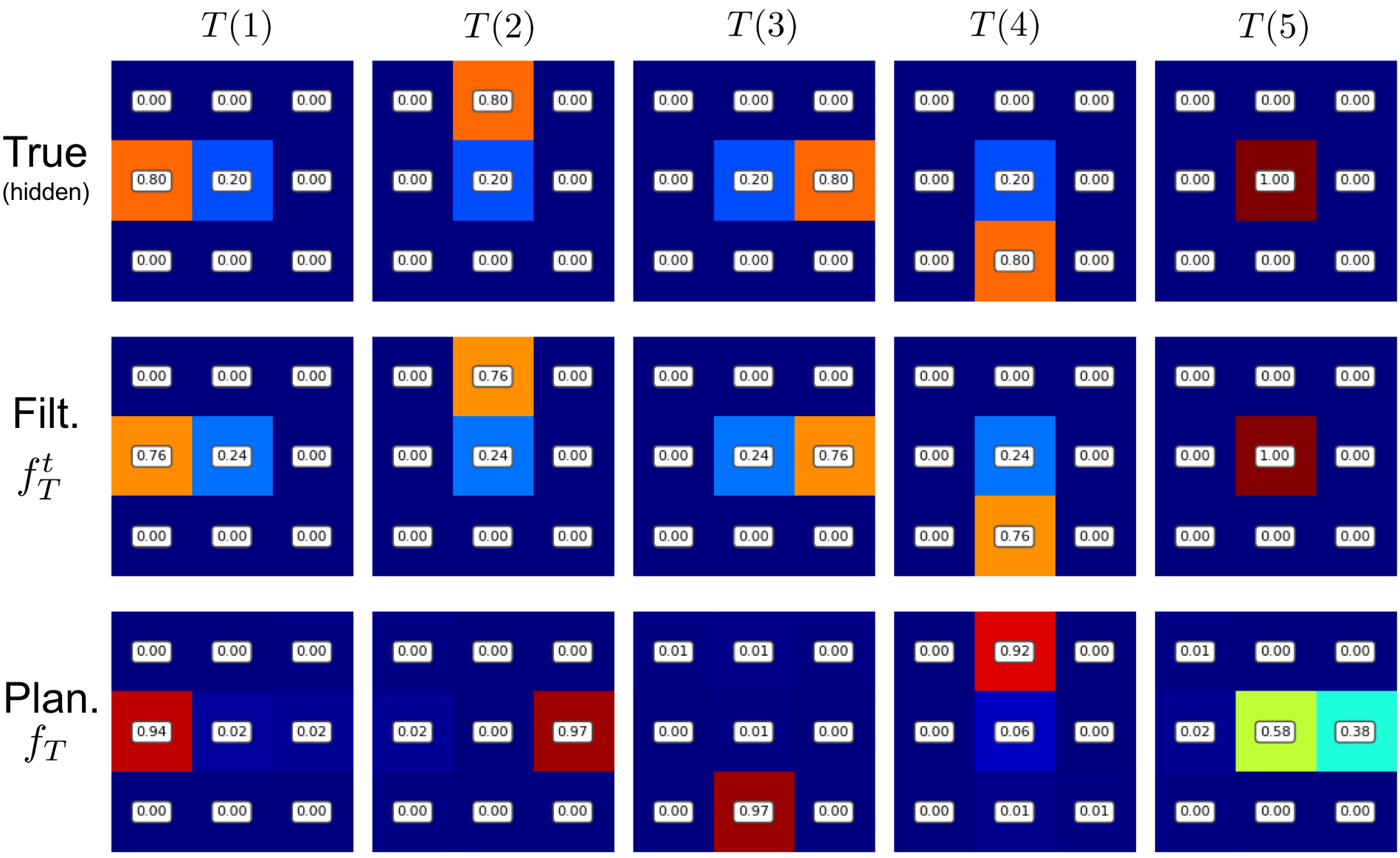}
  \caption{ Learned transition function $T$ in the  $\dense 18 \times 18$ stochastic grid navigation task.}
\label{fig:app_T}
\end{figure}

While both $f_{\sss T}$ and $f_{\sss T}'$ represent the same underlying transition dynamics, the learned transition probabilities are different in the filter and planner. Different weights allows each module to choose its own approximation and thus provides greater flexibility. The actions in the model $\modelspace a \in \modelspace A$ are learned abstractions of the agent's actions $\taskspace a \in \taskspace A$. Indeed, in the planner the learned transition probabilities for action $a_i \in A$ do not match the transition probabilities of  $\taskspace a_i \in \taskspace A$.

\subsection{Reward Function} 
Next plot the learned reward function $R$ for each action $a \in A$. 

\floatsetup[figure]{style=plain,subcapbesideposition=top}
\begin{figure}[H]
  \centering
	\includegraphics[width=0.98\textwidth]{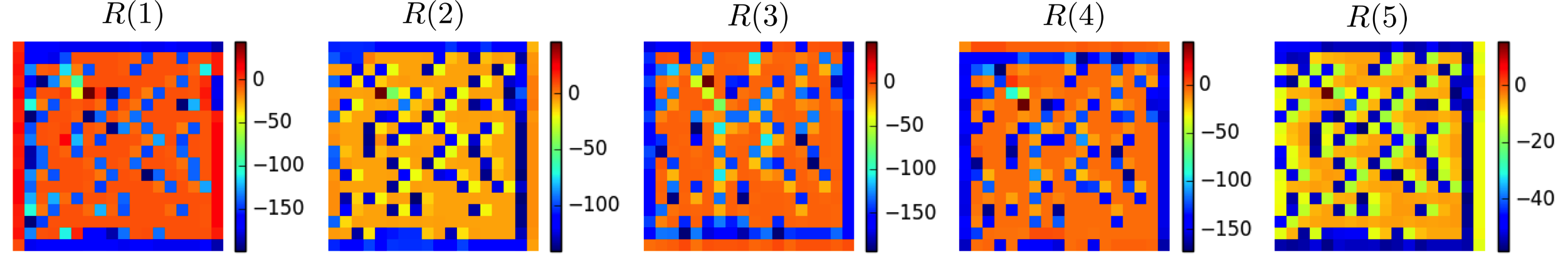}
  \caption{
Learned reward function $R$ in the  $\dense 18 \times 18$ stochastic grid navigation domain.
}
\label{fig:app_R}
\end{figure}

While the learned rewards do not directly correspond to rewards in the underlying task, they are reasonable: obstacles are assigned negative rewards and the goal is assigned a positive reward. Note that learned reward values correspond to the reward \emph{after} taking an action, therefore they should be interpreted together with the corresponding transition probabilities (third row of \figref{fig:app_T}).


\section{Implementation Details}\label{sec:app_details}
\subsection{Grid-World Navigation}\label{sec:app_gridnav}
We implement the grid navigation task in randomly generated discrete $N\mytimes N$ grids where each cell has $\dense p=0.25$ probability of being an obstacle. The robot has $5$ actions: move in the four canonical directions and stay put. Observations are four binary values corresponding to obstacles in the four neighboring cells. We consider a deterministic variant (denoted by -D) and a stochastic variant (denoted by -S). 
In the stochastic variant the robot fails to execute each action with probability $\dense P_t = 0.2$, in which case it stays in place. The observations are faulty with probability $\dense P_o = 0.1$ independently in each direction. Since we receive observations from $4$ directions, the probability of receiving the correct observation vector is only $0.9^4 = 0.656$. 
The task parameter, $\param$, is an $N\mytimes N\mytimes 3$ image that encodes information about the environment. The first channel encodes obstacles, $1$ for obstacles, $0$ for free space. The second channel encodes the goal, $1$ for the goal, $0$ otherwise. The third channel encodes the initial belief over robot states, each pixel value corresponds to the probability of the robot being in the corresponding state. 

We construct a ground-truth POMDP model to obtain expert trajectories for training. It is important to note that the learning agent has no access to the ground-truth POMDP models. In the ground-truth model the robot receives a reward of  $-0.1$ for each step, $+20$ for reaching the goal, and $-10$ for bumping into an obstacle. We use QMDP to solve the POMDP model, and execute the QMDP policy to obtain expert trajectories.
We use $10,000$ random grids for training. Initial and goal states are sampled from the free space uniformly. We exclude samples where there is no feasible path. The initial belief is uniform over a random fraction of the free space which includes the underlying initial state. More specifically, the number of non-zero values in the initial-belief are sampled from $\{1, 2, \dots N_f/2, N_f\}$ where $N_f$ is the number of free cells in the grid.
For each grid we generate $5$ expert trajectories with different initial state, initial belief and goal.
Note that we do not access the true beliefs after the first step nor the underlying states along the trajectory. 

We test on a set of $500$ environments generated separately in equal conditions. We declare failure after $10N$ steps without reaching the goal. Note that the expert policy is sub-optimal and it may fail to reach the goal.  We exclude these samples from the training set but include them in the test set.

We choose the structure of $\model(\param)$, the model in QMDP-net, to match the structure of the underlying task. The transition function in the filter $f_{\sss T}$ and the planner $f_{\sss T}'$ are both $3\mytimes3$ convolutions. While they both represent the same transition function we do not tie their weights. We apply a softmax function on the kernel matrix so its values sum to one. The reward function, $f_{\sss R}$, is a CNN with two convolutional layers. The first has $3\mytimes 3$ kernel, $150$ filters, ReLU activation. The second has $1\mytimes 1$ kernel, $5$ filters and linear activation. The observation model, $f_{\sss Z}$, is a similar two-layer CNN. The first convolution has a $3\mytimes 3$ kernel, $150$ filters, linear activation. The second has $1\mytimes 1$ kernel, $17$ filters and linear activation. The action mapping, $f_{\sss A}$, is a one-hot encoding function. The observation mapping, $f_{\sss O}$, is a fully connected network with one hidden layer with $17$ units and tanh activation. It has $17$ output units and softmax activation. The low-level policy function, $f_{\sss \pi}$, is a single softmax layer. The state space mapping function, $f_{\sss B}$, is the identity function. Finally, we choose the number of iterations in the planner module, $\dense K = \{30, 54, 90\}$ for grids of size $\dense N=\{10, 18, 30\}$ respectively.

The $3\mytimes 3$ convolutions in $f_{\sss T}$ and $f_{\sss Z}$ imply that $T$ and  $O$ are spatially invariant and local. In the underlying task the locality assumption holds but spatial invariance does not: transitions depend on the arrangement of obstacles. Nevertheless, the additional flexibility in the model allows QMDP-net to learn high-quality policies, \eg by shaping the rewards and the observation function. 

\subsection{Maze Navigation}
In the maze navigation task a differential drive robot has to navigate to a given goal. We generate random mazes on  $N \mytimes N$ grids using Kruskal's algorithm. The state space has $3$ dimensions where the third dimension represents $4$ possible orientations of the robot. The goal configuration is invariant to the orientation.  The robot now has $4$ actions: move forward, turn left, turn right and stay put.  The initial belief is chosen in a similar manner to the grid navigation case but in the 3-D space. The observations are identical to grid navigation but they are relative to the robot's orientation, which significantly increases the difficulty of state estimation. The stochastic variant (denoted by -S) has a motion and observation noise identical to the grid navigation. Training and test data is prepared identically as well. We use $\dense K = \{76, 116\}$ for mazes of size $\dense N=\{19, 29\}$ respectively.

We use a model in QMDP-net with a $3$-dimensional state space of size $N\mytimes N\mytimes 4$ and an action space with $4$ actions. The components of the network are chosen identically to the previous case, except that all CNN components operate on 3-D tensors of size $N \mytimes N \mytimes 4$. While it would be possible to use 3-D convolutions, we treat the third dimension as channels of a 2-D image instead, and use conventional 2-D convolutions. If the output of the last convolutional layer is of size $N \mytimes N \mytimes N_c$ for the grid navigation task, it is of size $N \mytimes N \mytimes 4N_c$ for the maze navigation task. When necessary, these tensors are transformed into a $4$ dimensional form $N \mytimes N \mytimes 4 \mytimes N_c$ and the max-pool or softmax activation is computed along the last dimension. 

\subsection{Object Grasping}
We consider a 2-D implementation of the grasping task based on the POMDP model proposed by \citet{hsiao2007grasping}. Hsiao et al. focused on the difficulty of planning with high uncertainty and solved manually designed POMDPs for single objects. 
We phrase the problem as a learning task where we have no access to a model and we do not know all objects in advance. In our setting the robot receives an image of the target object and a feasible grasp point, but it does not know its pose relative to the object. We aim to learn a policy on a set of object that generalizes to similar but unseen objects.

The object and the gripper are represented in a discrete grid. 
The workspace is a $14\mytimes 14$ grid, and the gripper is a ``U'' shape in the grid. The gripper moves in the four canonical directions, unless it reaches the boundaries of the workspace or it is touching the object. in which case it stays in place. The gripper fails to move with probability $0.2$. The gripper has two fingers with $3$ touch sensors on each finger. The touch sensors indicate contact with the object or reaching the limits of the workspace. The sensors produce an incorrect reading with probability $0.1$ independently for each sensor.  In each trial an object is placed on the bottom of the workspace at a random location. The initial gripper pose is unknown; the belief over possible states is uniform over a random fraction of the upper half of the workspace. 
The local observations, $o_t$, are readings from the touch sensors. The task parameter \param is an image with three channels. The first channel encodes the environment with an object; the second channel encodes the position of the target grasping point; the third channel encodes the initial belief over the gripper position. 

We have $30$ artificial objects of different sizes up to $6\mytimes 6$ grid cells. Each object has at least one cell on its top that the gripper can grasp. For training we use $20$ of the objects. We generate $500$ expert trajectories for each object in random configuration. We test the learned policies on $10$ new objects in $20$ random configurations each.
The expert trajectories are obtained by solving a ground-truth POMDP model by the QMDP algorithm. In the ground-truth POMDP the robot receives a reward of $1$ for reaching the grasp point and $0$ for every other state.  

In QMDP-net we choose a model with $S=14\times14$, $|A|=4$ and $|O|=16$. Note that the underlying task has $|\taskspace O| =64$ possible observations.  The network components are chosen similarly to the grid navigation task, but the first convolution kernel in $f_{\sss Z}$ is increased to  $5\mytimes 5$ to account for more distant observations. We set the number of iterations $\dense K=20$.

\subsection{Hallway2}
The Hallway2 navigation problem was proposed by \citet{littman1995learning} and has been used as a benchmark problem for POMDP planning~\cite{shani2013survey}. It was specifically designed to expose the weakness of the QMDP algorithm resulting from its myopic planning horizon. 
While QMDP-net embeds the QMDP algorithm, through end-to-end training QMDP-net was able to learn a model that is significantly more effective given the QMDP algorithm. 

Hallway2 is a particular instance of the maze problem that involves more complex dynamics and high noise. For details we refer to the original problem definition~\cite{littman1995learning}.
We train a QMDP-net on random $8\mytimes 8$ grids generated similarly to the grid navigation case, but using transitions that match the Hallway2 POMDP model. We then execute the learned policy on a particularly difficult instance of this problem that embeds the Hallway2 layout in a $8\mytimes 8$ grid. The initial state is uniform over the full state space. In each trial the robot starts from a random underlying state. The trial is deemed unsuccessful after $251$ steps.

\subsection{Navigation on a Large LIDAR Map}
\label{sec:app_real}
\begin{figure}[!ht]
  \centering
	 \includegraphics[height=4cm]{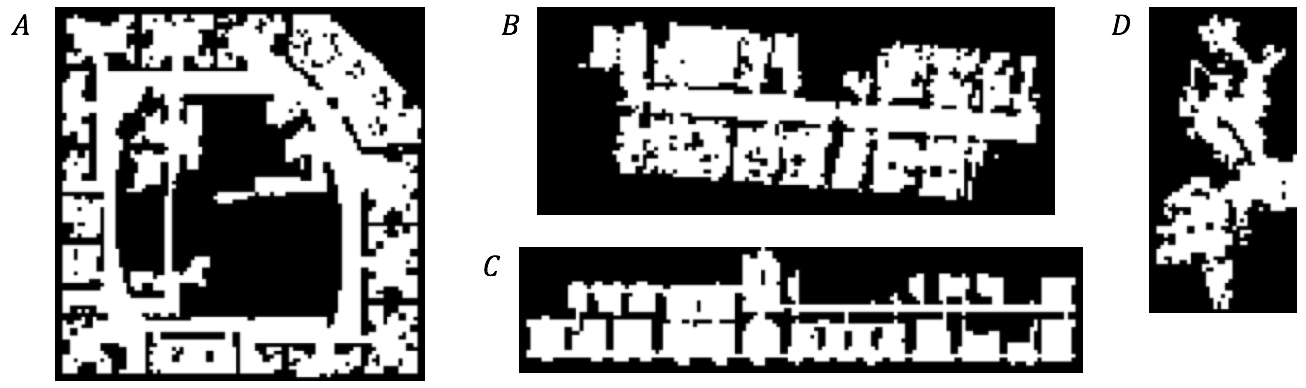}
  \caption{
Preprocessed $N \mytimes M$ maps. \textbf{A,} Intel Research Lab, $100 \mytimes 101$.  \textbf{B,} Freiburg, building 079, $139 \mytimes 57$.   \textbf{C,}  Belgioioso Castle, $151 \mytimes 35$. \textbf{D,} western wing of the MIT CSAIL building, $41 \mytimes 83$.
}
\label{fig:app_buildings}
\end{figure}

We obtain real-world building layouts using 2-D laser data from the Robotics Data Set Repository~\cite{radish}. More specifically, we use SLAM maps preprocessed to gray-scale images available online~\cite{unibonn}. 
We downscale the raw images to $NxM$ and classify each pixel to be free or an obstacle by simple thresholding. The resulting maps are shown in~\figref{fig:app_buildings}. We execute policies in simulation where a grid is defined by the preprocessed map. The simulation employs the same dynamics as the grid navigation domain. The initial state and initial belief are chosen identically to the grid navigation case. 

\floatsetup[figure]{style=plain,subcapbesideposition=top}

A QMDP-net policy is trained on the $30x30$-D grid navigation task on randomly generated environments. For training we set $K=90$ in the QMDP-net. We then execute the learned policy on the LIDAR maps. To account for the larger grid size we increase the number of iterations to $K=450$ when executing the policy.

\subsection{Architectures for Comparison}\label{sec:app_baselines}

We compare QMDP-net with two of its variants where we remove some of the POMDP priors embedded in the network (Untied QMDP-net, LSTM QMDP-net). We also compare with   two generic network architectures that do not embed structural priors for decision making (CNN+LSTM, RNN). We also considered additional architectures for comparison, including networks with GRU~\cite{cho2014learning} and ConvLSTM~\cite{xingjian2015convolutional} cells. ConvLSTM is a variant of LSTM where the fully connected layers are replaced by convolutions. These architectures performed worse than CNN+LSTM for most of our task.

\paragraph{Untied QMDP-net.} We obtain Untied QMDP-net by untying the kernel weights in the convolutional layers that implement value iteration in the planner module of QMDP-net. We also remove the softmax activation on the kernel weights. This is equivalent to allowing a different transition model at each iteration of value iteration, and allowing transition probabilities that do not sum to one.  In principle, Untied QMDP-net can represent the same policy as QMDP-net and it has some additional flexibility. However, Untied QMDP-net has more parameters to learn as $K$ increases. The training difficulty increases with more parameters, especially on complex domains or when training with small amount of data.

\paragraph{LSTM QMDP-net.} In LSTM QMDP-net we replace the filter module of QMDP-net with a generic LSTM network but keep the value iteration implementation in the planner. The output of the LSTM component is a belief estimate which is input to the planner module of QMDP-net. We first process the task parameter input $\param$, an image encoding the environment and goal, by a CNN. We separately process the action $\taskspace a_t$ and observation $\taskspace o_t$ input vectors by a two-layer fully connected component. These processed inputs are concatenated into a single vector which is the input of the LSTM layer. The size of the LSTM hidden state and output is chosen to match the number of states in the grid, \eg $N^2$ for an $\dense N \mytimes N$ grid.  We initialize the hidden state of the LSTM using the appropriate channel of the input $\param$ that encodes the initial belief.

\paragraph{CNN+LSTM.} CNN+LSTM is a state-of-the-art deep convolutional network with LSTM cells. It is similar in structure to DRQN~\cite{hausknecht2015deep}, which was used for learning to play partially observable Atari games in a reinforcement learning setting. Note that we  train the networks in an imitation learning setting using the same set of expert trajectories, and not using reinforcement learning, so the comparison with QMDP-net is fair.
The CNN+LSTM network has more structure to encode a decision making policy compared to a vanilla RNN, and it is also more tailored to our input representation. We process the image input, $\param$, by a CNN component and the vector input, $\taskspace a_t$ and $\taskspace o_t$, by a fully connected network component. The output of the CNN and the fully connected component are then combined into a single vector and fed to the LSTM layer. 

\paragraph{RNN.} The considered RNN architecture is a vanilla recurrent neural network with $512$ hidden units and tanh activation. At each step inputs are transformed into a single concatenated vector. The outputs are obtained by a fully connected layer with softmax activation.

We performed hyperparameter search on the number of layers and hidden units, and adjusted learning rate and batch size for all alternative networks. In particular, we ran trials for the deterministic grid navigation task. For each architecture we chose the best parametrization found. We then used the same parametrization for all tasks.

\subsection{Training Technique}\label{sec:training}
We train all networks, QMDP-net and alternatives, in an imitation learning setting. The loss is defined as the cross-entropy between predicted and demonstrated actions along the expert trajectories. We do not receive supervision on the underlying ground-truth POMDP models. 

We train the networks with backpropagation through time on mini-batches of $100$. The networks are implemented in Tensorflow~\cite{tensorflow2015-whitepaper}.  We use RMSProp optimizer~\cite{tieleman2012lecture} with $0.9$ decay rate and $0$ momentum setting. The learning rate was set to $1 \times 10^{-3}$ for QMDP-net and $1 \times 10^{-4}$ for the alternative networks. We limit the number of backpropagation steps  to $4$ for QMDP-net and its untied variant; and to $6$ for the other alternatives, which gave slightly better results.
We used a combination of early stopping with patience and exponential learning rate decay of $0.9$. In particular, we started to decrease the learning rate if the prediction error did not decrease for $30$ consecutive epochs on a validation set, $10\%$ of the training data. We performed $15$ iterations of learning rate decay. 

We perform multiple rounds of the training method described above. In our partially observable domains predictions are increasingly difficult along a trajectory, as they require multiple steps of filtering, \ie integrating information from a long sequence of observations. Therefore, for the first round of training we limit the number of steps along the expert trajectories, for training both QMDP-net and its alternatives. After convergence we perform a second round of training on the full length trajectories. 
Let  $L_r$ be the number of steps along the expert trajectories for training round $r$. We used two training rounds with $L_1=4$ and $L_2=100$ for training QMDP-net and its untied variant. For training the other alternative networks we used $L_1=6$ and $L_2=100$, which gave better results. 

We trained policies for the grid navigation task when the grid is fixed, only the initial state and goal vary. In this variant we found that a low $L_r$ setting degrades the final performance for the alternative networks. We used a single training round with $L_1=100$ for this task.

\end{document}